\documentclass[letterpaper]{article} 
\usepackage{aaai25}  
\usepackage{times}  
\usepackage{helvet}  
\usepackage{courier}  
\usepackage[hyphens]{url}  
\usepackage{graphicx} 
\usepackage{natbib}  
\usepackage{caption} 
\usepackage{algorithm}
\usepackage{algorithmic}
\usepackage{appendix}
\usepackage{booktabs}

\usepackage{amsfonts}
\usepackage{amsmath}

\usepackage{multirow}

\usepackage{wasysym}

\usepackage[hyperfootnotes=false, breaklinks=false]{hyperref}
\hypersetup{
    colorlinks=true, 
    linkcolor=red,  
    citecolor=blue,  
    urlcolor=blue,   
}
\usepackage[capitalise]{cleveref}

\usepackage{newfloat}
\usepackage{listings}
\DeclareCaptionStyle{ruled}{labelfont=normalfont,labelsep=colon,strut=off} 
\floatstyle{ruled}
\newfloat{listing}{tb}{lst}{}
\floatname{listing}{Listing}
\title{DiffGrasp: Whole-Body Grasping Synthesis Guided by Object Motion \\ Using a Diffusion Model}
\author{
    Yonghao Zhang\textsuperscript{\rm 1,\rm 2}\equalcontrib,
    Qiang He\textsuperscript{\rm 1,\rm 2}\equalcontrib,
    Yanguang Wan\textsuperscript{\rm 1,\rm 2},
    Yinda Zhang\textsuperscript{\rm 3},
    Xiaoming Deng\textsuperscript{\rm 1,\rm 2}\thanks{indicates corresponding author.},\\
    Cuixia Ma\textsuperscript{\rm 1,\rm 2\dag},
    Hongan Wang\textsuperscript{\rm 1,\rm 2}
}
\affiliations{
    \textsuperscript{\rm 1}Institute of Software, Chinese Academy of Sciences\\
    \textsuperscript{\rm 2}University of Chinese Academy of Sciences\\
    \textsuperscript{\rm 3}Google\\

    \{zhangyonghao2022, heqiang2022, wanyanguang2021, xiaoming, cuixia, hongan\}@iscas.ac.cn, \\
    yindaz@google.com
}

\usepackage{bibentry}

\begin{document}

\maketitle

\begin{abstract}
Generating high-quality whole-body human object interaction motion sequences is becoming increasingly important in various fields such as animation, VR/AR, and robotics. 
The main challenge of this task lies in determining the level of involvement of each hand given the complex shapes of objects in different sizes and their different motion trajectories, 
while ensuring strong grasping realism and guaranteeing the coordination of movement in all body parts. 
Contrasting with existing work, which either generates human interaction motion sequences without detailed hand grasping poses or only models a static grasping pose, we propose a simple yet effective framework that jointly models the relationship between the body, hands, and the given object motion sequences within a single diffusion model. To guide our network in perceiving the object's spatial position and learning more natural grasping poses, we introduce novel contact-aware losses and incorporate a data-driven, carefully designed guidance. 
Experimental results demonstrate that our approach outperforms the state-of-the-art method and generates plausible results.
\end{abstract}

%
\begin{links}
    \link{Project Page}{https://iscas3dv.github.io/DiffGrasp/}
\end{links}

\renewcommand\thesection{\arabic{section}}
\setcounter{secnumdepth}{1}

\section{Introduction}
\label{sec:intro}

Capturing, synthesizing, and controlling human motion plays a key role in many areas such as animation, VR/AR, and robotics. However, the movement of the human body is complex, especially the movement of hands. 
Hands are used in almost every scene of our daily life, from interacting with small objects like playing with toys or picking up and inspecting a phone, to handling larger objects like moving boxes or using the laptop with both hands. Manipulating different objects also necessitates different involvement of both hands, and controlling the motion of two hands is incredibly complex and nuanced. Realistically modeling such intricate interactions can greatly benefit downstream applications.

Existing work on human object interaction has made significant progress in modeling whole-body motion sequences, leveraging natural language \cite{diller2024cg, song2024hoianimator, ghosh2022imos}, object motion keypoints~\cite{li2023object, li2023controllable, taheri2022goal}, and initial frames \cite{interdiff, kulkarni2023nifty} to synthesize whole-body motion sequences or collaborative motion sequences. However, these works are not capable of generating satisfactory grasp sequences, especially for small objects because of the more intricate shapes of the object and the more flexible manipulations involved. In the domain of fine-grained grasping, GRIP \cite{taheri2023grip} and COOP \cite{COOP} can generate natural single frame whole-body grasping results based on the 3D positions of different small objects. However, these works are unable to generate realistic natural sequence results due to the lack of temporal smoothness modeling. 

\begin{figure}[!t]
    \centering
    \includegraphics[width=\columnwidth]{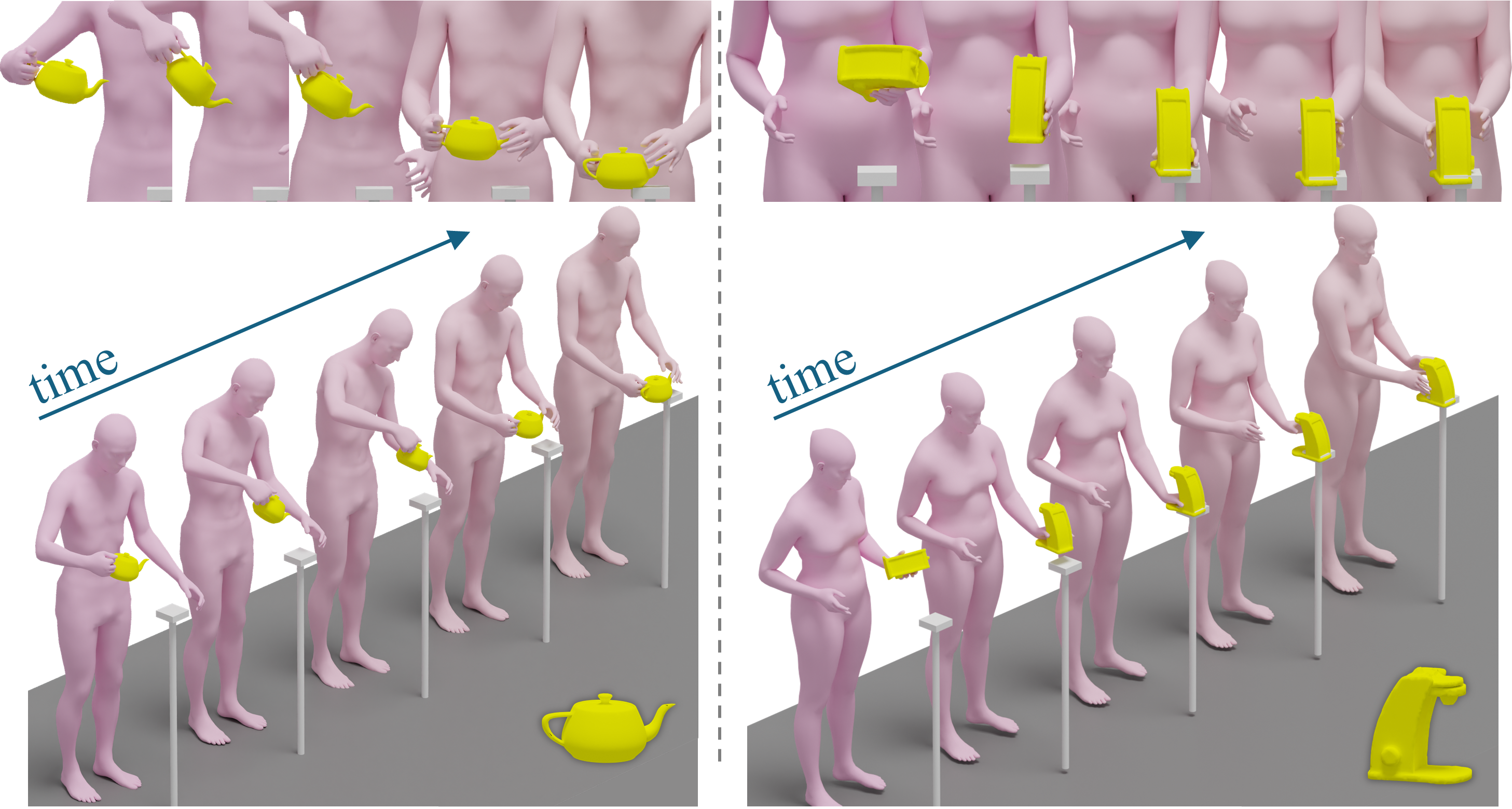}
    \caption{DiffGrasp generates whole-body human grasp sequence with realistic finger-object contact, conditioned on 3D object shape and object motion sequence. }
    \label{fig:cover}
\end{figure}

For generating realistic whole-body human object grasp motion sequences, there are three key challenges: first, the range of motion of the body and the range of motion of the hands are at two different spatial scales in speed and fineness, resulting in great difficulty to accurately model both motions together within a single model; 

second, when grasping objects of different sizes,  shapes and motions, the strategies for using either a single hand or both hands can be distinct due to complex grasp pattern with twice degree-of-freedom of both hands; finally, once an object is grasped, the contacting hand must maintain stable contact and avoid penetrating the object.

In this work, we propose DiffGrasp, a novel whole-body framework for generating fine-grained grasping sequences with both hands conditioned on an object motion sequence. A natural result of human interaction needs to consider both the shape of the object and the movement of the object to determine the involvement of both hands in grasping. 
Our method can generate natural, stable, and realistic whole-body grasping results for different object shapes while also maintaining coordinated movements of other body parts.
To model the complex movements of the object, hands and body jointly, we employ a novel conditional diffusion model to learn the joint distribution of the human-object motion space. 
This also avoids the network degradation issue that we observed in our experiments with existing methods, which may occur due to stacking of multiple diffusion stages.
To guide the network in learning complex motion patterns of the human body due to the high degrees-of-freedom in body with fingers, we propose two novel contact-aware losses, specifically for the hands, for training. The network can perceive the position of the object and generate realistic grasping results.
Finally, to enhance the realism of grasping, we introduce a data-driven guidance term during the inference stage to maintain contact stabilization, while other terms can encourage contact and prevent penetration.

The main contribution of our method can be summarized as follows. 
1) To the best of our knowledge, we propose the first diffusion-based framework to synthesize life-like whole-body human motion sequence with realistic finger-object contact, conditioned on 3D objects motion sequence.
2) We use a single diffusion model to learn the motion patterns between the hand and the object, while proposing two contact-aware losses that effectively guide the network to generate natural results that are also aware of the object's spatial position. 
3) We propose a novel data-driven guidance strategy to prevent sliding between the grasping hand and the object, while other guidance strategies can achieve more stable contact and avoid penetration of the object surface. 
4) Extensive experiments demonstrate that our proposed method outperforms the state-of-the-art method.

\section{Related Work}

\subsubsection{Human Motion Generation. }
Generating human-like motions is a fundamental problem of artificial intelligence and has gained significant attention in recent years. Previous studies have demonstrated the effectiveness of the Variational Autoencoder (VAE) formulation in generating diverse human motions \cite{petrovich2022temos, guo2022tm2t, lee2023multiact, lucas2022posegpt}. More recent work \cite{liu2024towards, zhang2024motiondiffuse, chen2023executing, ao2023gesturediffuclip, tseng2023edge, li2023finedance} has employed the diffusion model to generate motion sequences conditioned on various control signals, such as text, speech, music, etc. However, these control signals are generally ambiguous and the final motion sequences may not be well aligned with the user's real intention.
In addition, several recent studies \cite{pinyoanuntapong2024mmm, li2023finedance, lu2023humantomato} have explored the synthesis of whole-body human motion. These studies have typically adopted an independent modeling approach for hand parts and body parts, with the resulting outputs combined in a fusion step. 

\subsubsection{Hand Grasp Generation. }
There has been in-depth research on the problem of grasping with respect to given object poses, in the field of robotics \cite{sahbani2012overview}. 
With the emergence of human-object and hand-object interaction datasets \cite{zhang2021manipnet, fan2023arctic, H2o_2021_ICCV, GRAB:2020}, numerous works have arisen to generate single-frame hand grasp poses with contact map priors of objects \cite{jiang2021handobject, li2023contact2grasp, liu2023contactgen}.
For hand grasp sequence generation, D-Grasp \cite{christen2022dgrasp} synthesizes diverse dynamic sequences with the in-hand objects. Text2HOI \cite{cha2024text2hoi} performs this task by first generating the contact map and then the sequence of hand-object interaction. ArtiGrasp \cite{zhang2024artigrasp} can generate physically plausible bi-manual grasping.
In terms of modeling the hand-object interaction for better modeling a reasonable hand grasping pose, some approaches \cite{zhou2022toch,zheng2023cams,luo2024physics, liu2024geneoh} are proposed. These works achieve good results in hand grasp generation, but do not tackle whole-body grasp generation tasks. The generated hands positions also may not necessarily produce natural poses for the body.

\subsubsection{Human Object Interaction Generation. }
Existing works on generating interactions between human body and objects can be categorized based on the given object. 
Interacting with scene objects \cite{cen2024text_scene_motion, kulkarni2023nifty, huang2023diffusion, zhao2023synthesizing} refers to interact with objects in the environment that cannot be moved by a person. 
In the realm of interactions with larger objects \cite{song2024hoianimator, diller2024cg, Li_2024_WACV, li2023object, li2023controllable, interdiff, peng2023hoi}, most of these related works focus on reconstructing well-defined torso motion sequences with text guidance often provided as a prompt \cite{peng2023hoi,li2023controllable}, while overlooking hand poses. 
For tasks involving interactions with smaller objects, existing works usually generate fine-grained single-frame grasping poses \cite{taheri2022goal,wu2022saga,tendulkar2023flex,COOP}, while sequence generation tends to adopt methods by which hand motion guides object motion to achieve perceptually plausible results \cite{ghosh2022imos, interdiff, Li_2024_WACV}.
In summary, previous work has not been able to generate realistic whole-body human motion sequences while maintaining a fine hand-object grasping posture.

\begin{figure*}[t]
    \centering
    \includegraphics[width=0.9\textwidth]{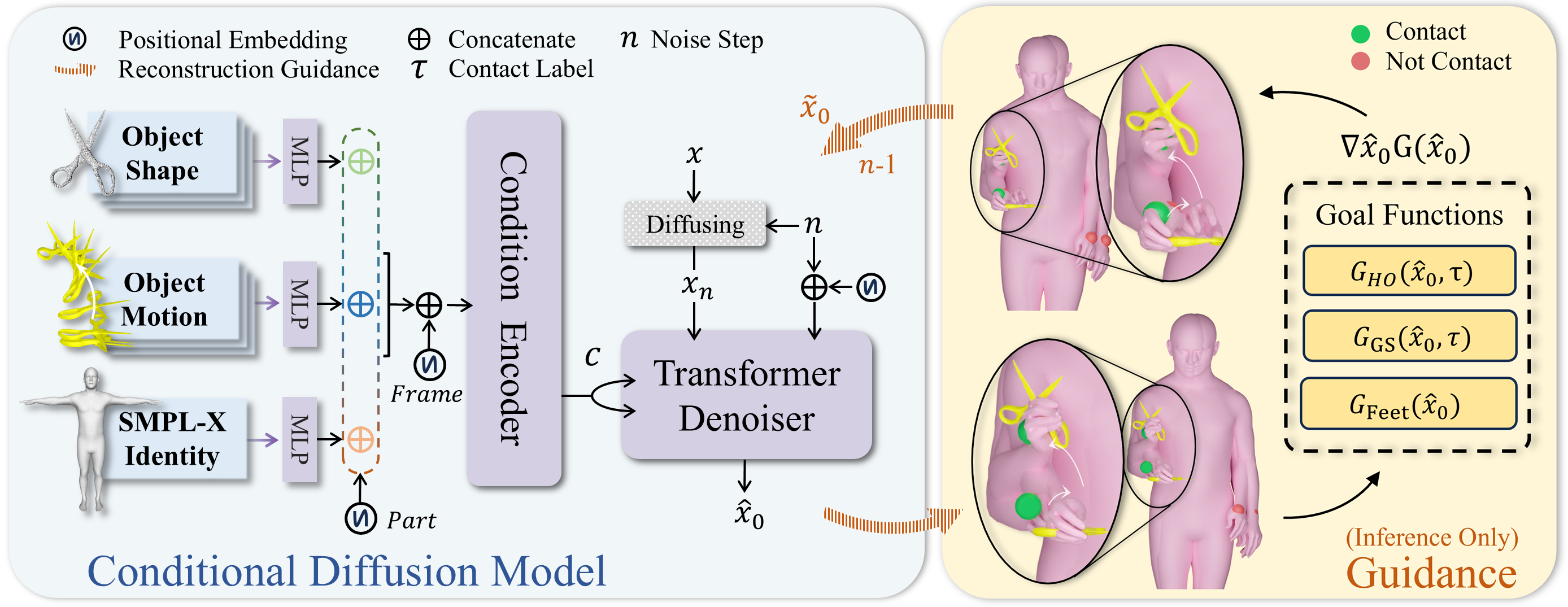}
    \caption{\textit{Overview of DiffGrasp Framework.} In our conditional diffusion model, we use the given sequence of object motion, object shape and the SMPL-X identity as conditions. After specially designed positional encodings, these embedded conditions are inputted into a transformer-encoder-based condition encoder. Then, a transformer decoder as denoising network predicts a sequence of clean whole-body pose of SMPL-X as well as the wrist joints translations relative to the object centroid. During the inference stage, we reconstruct the SMPL-X pose sequence into a human mesh sequence. Based on carefully designed guidance functions, we control and optimize our predicted results for more stable hand grasping ($\mathcal{G}_{GS}$), less penetration ($\mathcal{G}_{HO}$) and better foot-floor contact ($\mathcal{G}_{Feet}$) through reconstruction guidance strategy.}
    \label{fig:big_pipe}
\end{figure*}

\section{Method}
DiffGrasp generates whole-body human motion sequence with realistic finger-object contact, mainly conditioned on 3D objects motion sequence and human identity. The overview of DiffGrasp is shown in \cref{fig:big_pipe}. 
We will introduce our model in three parts: data representation, diffusion model and contact-aware loss, and guidance.

\subsection{Data Representation}

\subsubsection{Human Motion Representation.}
We denote the generated human whole-body pose as $H\in\mathbb{R}^{T\times D}$, 
where $T$ and $D$ represent the sequence length and dimension of the human pose. 
In frame $t$, the human pose $H_t$ consists of global translation and global orientation of the root joint, and the body pose using the 6D continuous representation\cite{zhou2020continuity}. 
We use SMPL-X\cite{SMPL-X:2019} to represent the human body and use its shape $\beta$ and gender $G$ to present \textit{Human Identity} $S_{id}=\{\beta, G\}$.

\subsubsection{Object Sequence Representation.}
The object motion, which is one of our conditions, is represented by two components: the global 3D motion sequences for \textit{Object Motion} and the geometry information for \textit{Object Shape}. The motion of the object in each frame is represented by the centroid of the object and the global rotation and translation of the object, denoted as $W_t\in\mathbb{R}^{12}$. 
For the shape of the object, following previous work \cite{taheri2022goal, COOP, li2023object}, we represent the geometry of the object using the Basis Point Set (BPS) representation \cite{prokudin2019efficient}. In each frame $t$, we calculate the 3D mesh BPS representation of the object, denoted as $V_t\in\mathbb{R}^{1024\times 3}$. 

\subsubsection{Condition Input.}
At each frame $t$, we first use three MLPs to map each object shape $V_t$, object motion $W_t$ and human identity $S_{id}$ to three 256 dim features $\mathcal{V}_t,\mathcal{W}_t$ and $\mathcal{S}_{id}$. 
Inspired by \cite{cha2024text2hoi}, in addition to the general frame-wise positional encoding $PE^f_t$, we introduce the part-wise positional encoding $PE^p$ to provide a more detailed differentiation of the frames and the condition part. 
The final sequence of input conditions can be formulated as
$c^{raw}_t=PE^f_t(PE^p_V(\mathcal{V}_t)\oplus PE^p_W(\mathcal{W}_t)\oplus PE^p_s(\mathcal{S}_{id}))\in\mathbb{R}^{256\times3}$. 

\noindent More details of our positional encoding strategy can be found in the supplementary materials.

\subsection{Conditional Diffusion Model}

\subsubsection{Model Architecture.}
\cref{fig:big_pipe} shows our model architecture.
We adopted a basic full Transformer \cite{vaswani2017attention} architecture to fulfill our sequence-to-sequence generation task. Our network includes a transformer encoder as a condition encoder to encode the condition sequence $c^{raw}$ into condition features $c$, and a transformer decoder as a denoiser, which maps the noised motion sequence data $x_n$ and the noise step $n$ to predict the clean data $x_0$ conditioned on $c$. 

\subsubsection{Denoiser Outputs.}
The conditional diffusion model aims to learn the latent correspondence between input conditions $c$ and the generated whole-body human motion sequence $H$. 
To get more accurate 3D Euclidean space awareness for our guidance stage, 
our model also predicts translations of both hands wrist joint $\kappa\in\mathbb{R}^{T\times 6}$ relative to the condition object center $O_m$. The generated result is denoted as $X=\{H,\kappa\}$.

\subsubsection{Conditional Diffusion Loss.}
The diffusion model comprises both a forward (noising) process and a reverse (de-noising) process. The forward process involves gradually introducing noise to the initial data representation $x_0$ over $N$ steps, implemented through a Markov chain formulation,
\begin{gather}
    q(x_n | x_{n-1})=\mathcal{N}(x _n;\sqrt{1-\beta_n}x_{n-1},\beta_n I) \\
    q(x_{1:N}|x_0):=\sum^N_{n=1}q(x_n|x_{n-1})
\end{gather}

\noindent where $\beta_n$ is a fixed variance schedule and $I$ is an identity matrix. For our method, the goal is to learn a conditional diffusion model $f_\theta$ to reverse the noising diffusion process,
\begin{equation}
    f_\theta(x_{n-1}|x_n,c):=\mathcal{N}(x_{n-1};\mu_\theta(x_n,n,c),\Sigma_n)
    \label{eq:mu_theta}
\end{equation}
where $\mu_\theta$ denotes the predicted mean and $\Sigma_n$ is a fixed variance. The process of learning the mean $\mu_\theta$ can be optimized by reconstructing clean data $x_0$ following existing motion generation methods \cite{tevet2023human, shafir2023human}. The model $f_\theta$ is optimized by the objective:
\begin{equation}
    \mathcal{L}_{\textit{diff}}=\mathbb{E}_{n\sim[1,N]}\Vert f_\theta(x_n,n,c)-x_0 \Vert^2_2
\end{equation}

\subsubsection{Contact Label.}
To independently represent the contact relationship of each hand with the object, we design a binary hand-object mask $\tau\in\{0,1\}^{T\times2}$, named \textit{contact label}. The two components of $\tau_t$ are represented by 1 or 0, respectively, indicating whether the left hand or the right hand are in contact with the object at frame $t$. Contact label is calculated by thresholding the minimum distance between each hands and the mesh vertices of a given object geometry. During inference, the contact label is obtained by calculating whether the relative distance $\kappa$ between the generated wrist and object is less than a given threshold.

\subsubsection{Contact-aware Losses.} 
Since the SMPL-X pose representation is on SE(3), and modeling human interacting with objects requires explicit representation in the 3D Euclidean space, we additionally propose two contact-aware loss functions for training based on contact label $\tau$ to provide more clues in the 3D Euclidean space for our network.

\noindent\textit{Contact-aware Reconstruction Loss} enforces  the generated hand joints and wrist joints to be close to ground truth. 
After the network predicts $\hat{X}=\{\hat{H},\hat{\kappa}\}$, we use the pose parameters $\hat{H}$ to reconstruct all joints of the left hand and right hand $\hat{J}$ using the forward process of SMPL-X, and we denote left hand joints and right hand joints as $\hat{J_l}$, and $\hat{J_r}$, respectively. Furthermore, by adding the 3D position of the object centroid with the translation of the predicted wrist, we reconstruct the positions of the hand wrist joint, denoted as $\hat{\upsilon_l}$ and ${\hat{\upsilon_r}}$ by $\hat{\upsilon}=O_m+\hat{\kappa}$, where $O_m$ is the centroid of the object shape. Then, the contact-aware reconstruction loss can be formulated as:

\begin{equation}
    \begin{aligned}
    \mathcal{L}_{\textit{recon}}&=\tau_0(\Vert J_l-\hat{J}_l \Vert_2 + \lambda_{\textit{wrist}} \Vert v_l - \hat{v}_l \Vert_2) \\
    &+\tau_1(\Vert J_r-\hat{J}_r \Vert_2 + \lambda_{\textit{wrist}} \Vert v_r - \hat{v}_r \Vert_2)
    \end{aligned}
\end{equation}
where $\tau^0$ and $\tau^1$ are the contact label of left and right hands, respectively, and $\lambda_{\textit{wrist}}$ is the balance weight of wrist terms.

\noindent \textit{Contact-aware Interaction Loss} can ensure an accurate spatial position of the hands with respect to the object. In order to make the model more sensitive to the proximity of both hands to the object, we introduce exponentially decaying distance-aware interaction weights $w_k$ for each hand joints $k$, inspired by \cite{ghosh2023remos}:

\begin{equation}
    w_k = \tau \exp{(-\alpha \cdot d(J_k,O_m))}
\end{equation}
where $d(\cdot, \cdot)$ represents the per-joint Euclidean distance, and $\alpha$ is a scalar weight. Thus, the contact-aware interaction loss is defined as:
\begin{equation}
    \mathcal{L}_{\textit{inter}}=\sum^K_{k=1}w_k \Vert d(\hat{J_k}, O_m)-d(J_k,O_m)) \Vert_2
\end{equation}

In total, we train our DiffGrasp to minimize a weighted sum of three loss terms: the diffusion loss, the contact-aware reconstruction loss and the contact-aware interaction loss:
\begin{equation}
    \mathcal{L} = \lambda_{\textit{diff}}\mathcal{L}_{\textit{diff}} + 
    \lambda_{\textit{recon}} \mathcal{L}_{\textit{recon}} + 
    \lambda_{\textit{inter}} \mathcal{L}_{\textit{inter}}
\end{equation}
where $\lambda_{\textit{diff}}$ and $\lambda_{\textit{contact}}$ are assigned scalar weights to balance the individual losses.

\subsection{Guidance}

In human object interaction field, direct constraints between the human body mesh and the object mesh are crucial for generating realistic results. However, during the training stage, we do not explicitly impose loss constraints between the human body and the object meshes. This is because we found that, first, DiffGrasp, constrained by the aforementioned losses, can already generate reliable results; second, introducing explicit penetration and contact losses during training would significantly increase computational costs and training time.
However, these factors do not significantly improve model generalization and visual realism. 
Therefore, we only explicitly optimize the interaction during the inference sampling process, because it would better balance the constraints and enhance visual realism by fitting the observed object motions.

\begin{figure}[t]
    \centering
    \includegraphics[width=\columnwidth]{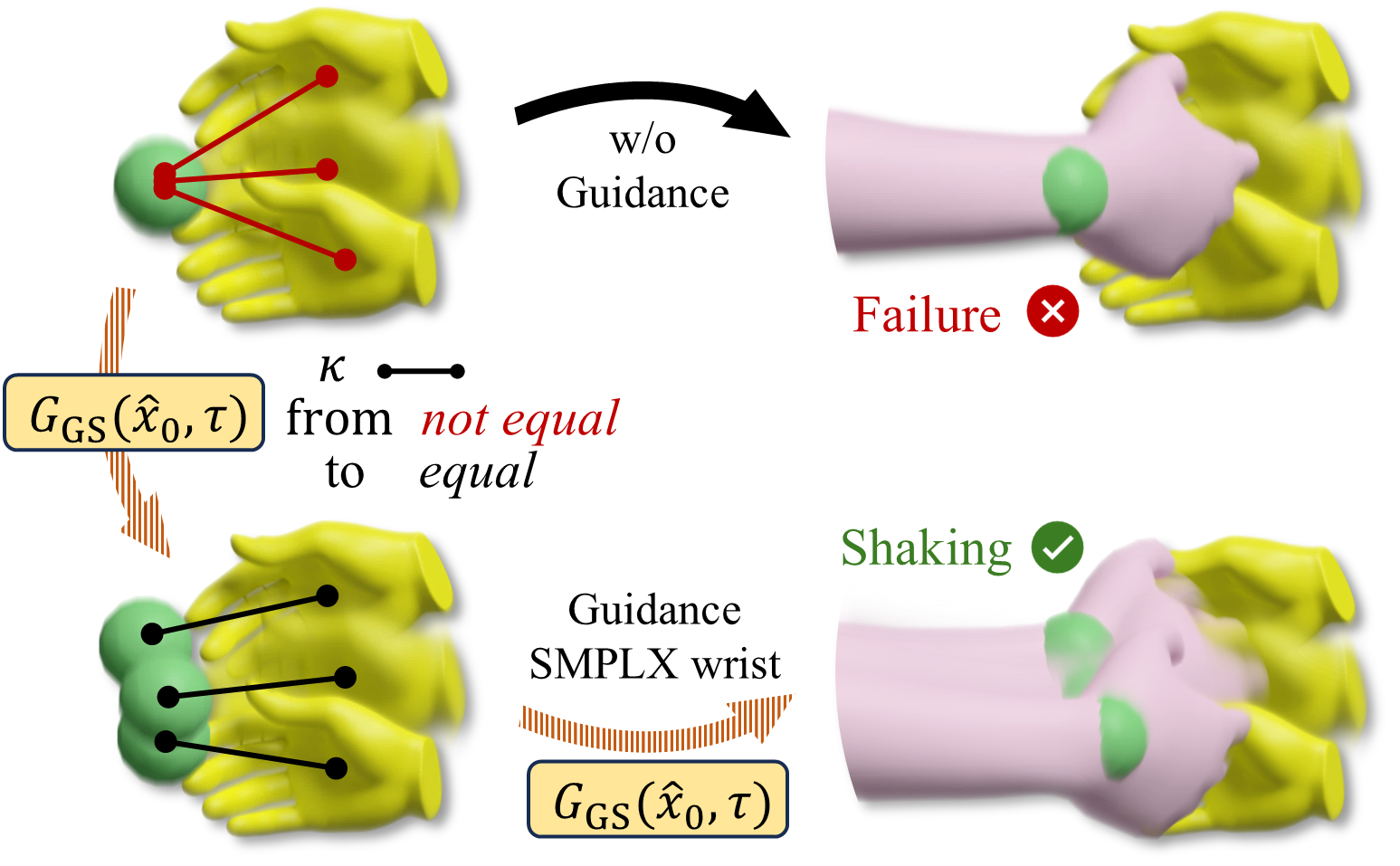}
    \caption{\textit{Illustration of Grasp Stabilization Guidance $\mathcal{G}_{GS}$.} 'Handshaking' object movement example: Initially, the generated hand-object relative distance $\kappa$ and the reconstructed hand do not follow the object's (yellow hand) shaking well. We stabilized the hand-object relative distance according to \cref{eqa:10} to obtain the wrist position that follows the object's shaking, and then guided the reconstructed wrist position to successfully achieve the handshaking effect.}
    \label{fig:guidance}
\end{figure}

\subsubsection{Reconstruction Guidance.}
To align the model's inference sampling results more with a specific condition, the diffusion model often uses an optimization strategy called classifier guidance \cite{ho2020denoising}. This strategy can be formulated as $ \Tilde{\mu} = \mu - \eta\Sigma_n\nabla_{\mu} \mathcal{G}(\mu)$, where $\mu$ is the result of the denoised step $n$ defined by \cref{eq:mu_theta}, $\mathcal{G}$ is a goal function that determines a gradient direction that aligns better with the condition, and $\eta$ is a scalar weight. 
Similarly, we optimize the reconstructed results, known as reconstruction guidance \cite{ho2022video}, which can be formulated as,
\begin{equation}
    \Tilde{x}_0 = \hat{x}_0 - \eta\Sigma_n\nabla_{\hat{x}_n} \mathcal{G}(\hat{x}_0)
\end{equation}
where $x_0$ represents the clean data predicted by the network, in our work it corresponds to the SMPL-X human parameters $\hat{H}$ and the relative object translation of two wrists $\hat{J}$. We employ reconstruction guidance to gradually guide the optimization of the predicted results by the network. Through reconstructing $\hat{H}$, we obtain the human body mesh surface. We will detail our goal function $\mathcal{G}$ in the following.

\subsubsection{Grasp Stabilization Guidance.}
Instead of hand-crafted constraints, we use data-driven constraints generated
from our model. We introduce the Grasp Stabilization Guidance, which is motivated by a simple observation: \textit{when a hand successfully grasps an object, it should not exhibit any relative sliding motion with respect to the object.} This guidance is especially helpful in optimizing the generated results for high-frequency, rapid object movements. Without loss of generality, we consider the generated relative distance $\hat{\kappa}_l$ between the left wrist and the object as an example.

Given an sequence of left wrist relative distance $\hat{\kappa}_l^0, \hat{\kappa}_l^1,\ldots, \hat{\kappa}_l^n$, its sequence of contact label $\tau_0^0, \tau_0^1, \ldots, \tau_0^n$ obtained by calculating whether the relative distance $\hat{\kappa}_l$ is less than a specific threshold, assuming the temporal segment of contact is from $i$ to $j$, where $0 \leq i < j \leq n$, our aim is to ensure that the position of the wrist relative to the object in the $i$-th frame does not change in the next $j-i$ frame. For any frame $k$ in the next $j-i$ frame, the corrected wrist position in the world coordinate system is calculated by the following formula:    
\begin{equation}
    \acute{v}_l^k=(\hat{\kappa}_l^i+O_m-W_T^i){W_R^i}^{\text{T}}W_R^k+W_T^k
    \label{eqa:10}
\end{equation}
where $W_R^k$ and $W_T^k$ is the object rotation and translation at frame $k$. We demonstrate this correction in \cref{fig:guidance}. Next, we use $\hat{H}$ to reconstruct human wrist joints $\hat{v}$, and then constrain them to optimized wrist joints. Grasp stabilization guidance can be defined as
\begin{gather}
    \mathcal{G}_{GS} =  \tau \Vert \hat{v} - \acute{v} \Vert_2  \\
    \Tilde{x}_0^{up} = \hat{x}_0^{up} - \eta\Sigma_n\nabla_{\hat{x}_n^{up}} \mathcal{G_{GS}}(\hat{x}_0, \tau)
\end{gather}

Using contact labels $\tau$, only the contact hand needs to be optimized. We optimize only the upper body parameters $\Tilde{x}_0^{up}$ in experiments.

\subsubsection{Hand-Object Contact Guidance.}
To reduce the penetration between the reconstructed human mesh and the object mesh, and to encourage contact between the hands and the object, we propose hand-object contact guidance. To expedite the computation process, we followed the work \cite{hasson2019learning} to sample only $|v_s|$ vertices on hands with the highest contact rates for penetration and contact distance calculations. Penetration distance $D_{pene}$ and contact distance $D_{cont}$ can be calculated as:
\begin{align}
    D_{pene} &= \sum\nolimits_{h=1,\ldots,\hat{v}_s}^{}{-\text{min}\{\text{sdf}(\hat{v}_s[h]),0\}} \\
    D_{cont} &= \sum\nolimits_{h=1,\ldots,\hat{v}_s}^{}{\tau \cdot \text{abs}(\text{sdf}(\hat{v}_s[h]))}
\end{align}

We compute the signed distance between the sampling points $v_s$ on the hand and the object represented with signed distance fields (SDF), and simultaneously calculate the penetration distance and the contact distance. Then, we can define the hand-object contact guidance as follows:
\begin{gather}
    \mathcal{G}_{\textit{HO}} =  \lambda_{\textit{ho}}D_{\textit{pene}} + (1-\lambda_{\textit{ho}})D_{\textit{cont}} \\
    \Tilde{x}_0^{\textit{hand}} = \hat{x}_0^{\textit{hand}} - \eta\Sigma_n\nabla_{\hat{x}_n^{\textit{hand}}} \mathcal{G}_{\textit{HO}}(\hat{x}_0,\tau)
\end{gather}

During optimization, only the contact hand can be optimized. We only optimize the parameters of the hands $\Tilde{x}_0^{\textit{hand}}$.

\subsubsection{Feet Penetration Guidance.}
We use feet penetration guidance that encourages human feet-floor contact:
\begin{gather}
    \mathcal{G}_{\textit{Feet}} =  \sum\nolimits_{\hat{v}[z]<0} \text{abs}(\hat{v}[z]) \\
    \Tilde{x}_0 = \hat{x}_0 - \eta\Sigma_n\nabla_{\hat{x}_n} \mathcal{G}_{\textit{Feet}}(\hat{x}_0)
\end{gather}

We use these three goal functions to explicitly guide the diffusion in generating results that better match the object motion condition. The goal function can be defined as:
\begin{equation}
    \mathcal{G} = \mathcal{G}_{\textit{Feet}} + \mathcal{G}_{\textit{GS}} + \mathcal{G}_{\textit{HO}}
\end{equation}

\section{Experiments}
In this section, we first describe the dataset and evaluation metrics used in our experiment. Next, we present comparison experiments with the baseline method. Finally, we conduct ablation studies to evaluate the effectiveness of our method.

\subsection{Datasets and Evaluation Metrics}

\subsubsection{Datasets.}
We use GRAB \cite{GRAB:2020} and ARCTIC \cite{fan2023arctic} to conduct our experiment, which collects full-body hand-object interaction mesh sequences. Each dataset consist of 10 subjects, each grabbing and manipulating a number of different objects. We follow the conventional approach to divide the training and validation sets by 8 subjects used for training and 2 subjects for testing. To further evaluate our model’s generalization ability to unseen objects, we exclude five objects from the GRAB training dataset for testing. In ARCTIC, we use the ground truth of the articulation angles of objects.

\subsubsection{Evaluation Metrics.} 

We use a set of evaluation metrics to this novel task, building upon established evaluation methods in the existing literature \cite{li2023object, taheri2022goal, li2023controllable,he2022nemf}. 

\begin{table*}[!hbpt]
    \centering
    \fontsize{9}{11}\selectfont
 \begin{tabular}{l|l|cccccccc}
    
\toprule											
Dataset & Method	&	Hands JPE$\downarrow$	&	 MPJPE$\downarrow$	&	 MPVPE$\downarrow$	&	 FS$\downarrow$	&	 Coll. \%$\downarrow$	&	 C depth $\downarrow$ 	&	 F1$\uparrow$ 	&	 Cont dist$\downarrow$ \\
\midrule											

\multirow{4}{*}{GRAB} & OMOMO	&	31.28 	&	17.57 	&	13.80 	&	1.05  	&	0.0014 	&	0.0007 	&	0.0090 	&	0.19\\
 & OMOMO-V2 	& 	34.91 	&	19.47 	&	15.31 	&	2.63 	&	0.0007 	&	0.0009 	&	0.0641 	&	0.25\\
 & OMOMO-V3	& 	32.72 	&	17.45 	&	13.75 	&	\textbf{1.03} 	&	0.0004 	&	0.0001 	&	0.1028 	&	0.15 \\

\cmidrule{2-10}

 & \textbf{DiffGrasp}	& 	\textbf{20.99} 	&	\textbf{12.24} 	&	\textbf{10.09} 	&	2.22 	&	0.0023 	&	0.0001 	&	\textbf{0.7840} 	&	\textbf{0.04} 	\\		

\midrule

\multirow{4}{*}{ARCTIC} & OMOMO	&	25.95 	&	11.68 	&	8.53 	&	\textbf{0.64} 	&	0.0001 	&	0.0001 	&	0.0775 	&	0.11\\
                        & OMOMO-V2 	& 	26.91 	&	11.77 	&	8.50 	&	0.93 	&	\textbf{0.0000} 	&	\textbf{0.0000} 	&	0.2771 	&	0.12\\
                        & OMOMO-V3	& 	26.57 	&	11.91 	&	8.74 	&	0.65 	&	\textbf{0.0000} 	&	0.0001 	&	0.3338 	&	0.10 \\

\cmidrule{2-10}

 & \textbf{DiffGrasp}	& 	\textbf{19.96} 	&	\textbf{11.56} 	&	\textbf{8.00} 	&	1.25 	&	0.0030 	&	0.0001 	&	\textbf{0.8067} 	&	\textbf{0.04} 	\\	

\bottomrule			

    \end{tabular}
    \caption{\textit{Comparative experimental results }on GRAB \cite{GRAB:2020} dataset and ARCTIC \cite{fan2023arctic} dataset. }
    
    \label{tab:experimental}
\end{table*}

\noindent \paragraph{\textit{Motion Quality Metrics.}} \textit{HandJPE} and \textit{MPJPE} represent mean hand joint position errors (cm) and mean per-joint position errors (cm), respectively. \textit{MPVPE} represents mean per-vertex errors (cm), 
and \textit{FS} represents foot sliding metric.

\noindent \paragraph{\textit{Hand Collision Metrics.}} \textit{Collision Percentage} and \textit{Collision Depth} are used to evaluate the extent to which the vertices of the hands penetrate the surface of an object. At frame $t$, we employ the signed distance field of object to judge whether a hand vertex is located within the mesh and to calculate the distance between the vertex and the mesh surface. If the vertex is inside the mesh and the distance is below a specified threshold (5mm), we increase the collision count and record the collision depth. By iterating through the sequence, we can compute the collision percentage and the mean collision depth.

\noindent \paragraph{\textit{Hand Contact Metrics.}} In order to evaluate the effectiveness of hand contact, we employ the
\textit{F1 score} metric commonly used in object detection tasks. To obtain a more accurate measure of grasping, we calculate the mean \textit{Contact distance} (cm) between the positions of the fingers and the object meshes. We empirically define a contact threshold of 5mm and use it to determine the contact labels for each frame. The same calculation is performed for the ground truth hand positions. Subsequently, we tally the true/false positive/negative cases to compute the F1 score.

\subsection{Evaluations}
\subsubsection{Baselines.} 

We compare our DiffGrasp with OMOMO \cite{li2023object}, a two-stage framework that generates whole-body motion sequences without the finger pose, conditioned on the object motion sequence. It employs two conditional diffusion models: the first generates hand positions conditioned on object geometry features, and the second generates whole-body poses based on the predicted hand positions. 
To facilitate a comprehensive comparison, we trained three versions of OMOMO based on its basic network architecture, namely OMOMO, OMOMO-V2, and OMOMO-V3. OMOMO follows the original settings, using the object motion trajectory and the BPS representation as the first-stage network inputs, which generates hand positions. These positions then condition the prediction of body pose in the second stage. OMOMO-V2 extends OMOMO by including the parameters of hand pose as additional outputs in its second model. 
OMOMO-V3 is built on the architecture of OMOMO, with an additional network that predicts hand pose parameters conditioned on the object motion trajectory, the BPS representation, and the hand positions. 
The additional network can be treated as predicting the hand pose parameters in its first hand positions generation model. 
Comparisons and results with more related work can be found in our supplementary materials.

\subsubsection{Results.}

\begin{figure}[!htbp]
    \centering
    \includegraphics[width=\columnwidth]{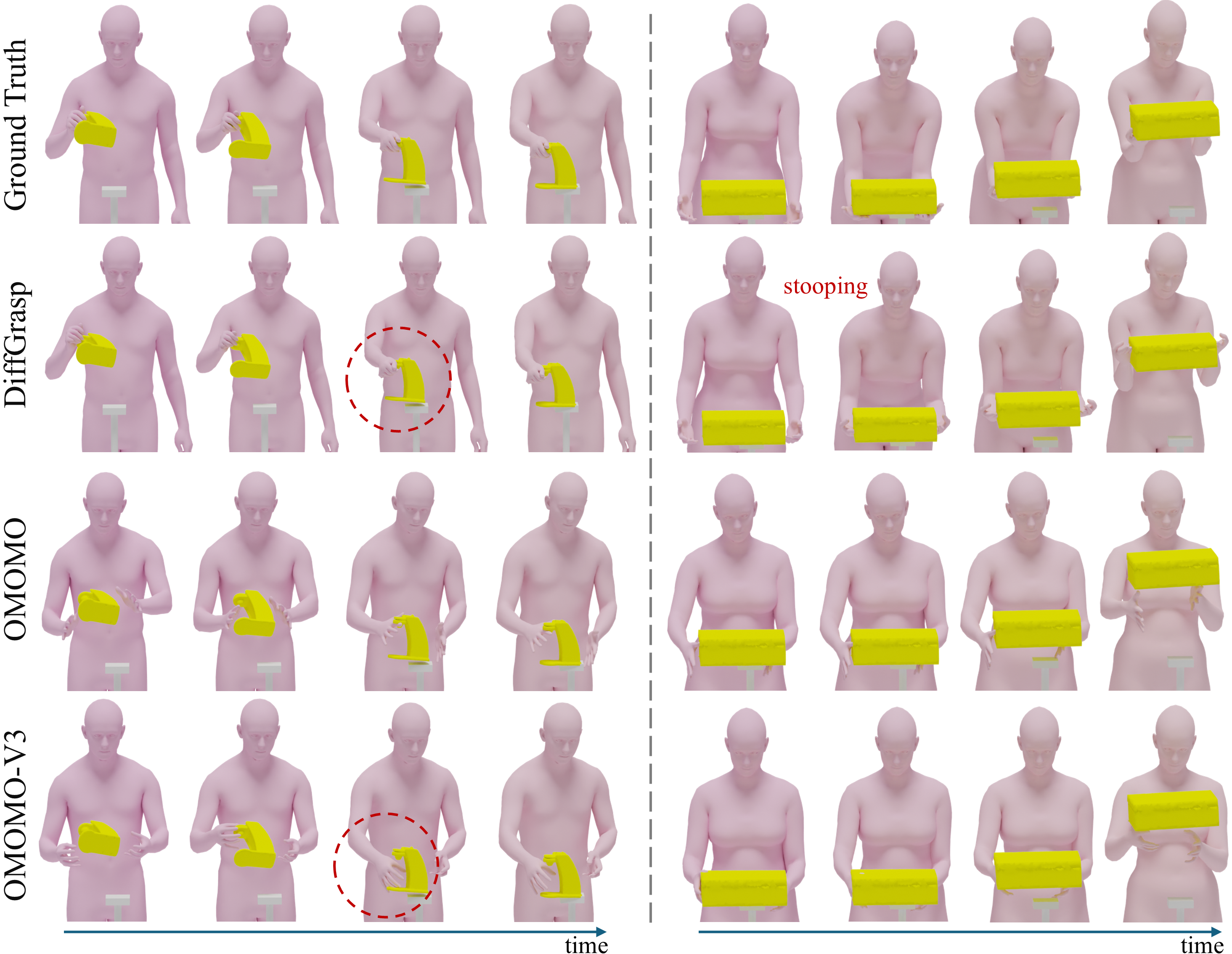}
    \caption{\textit{Qualitative Results of Comparison Experiments.} Our model (DiffGrasp) generates more realistic results, with more hand-object contact and less penetration. }
    \label{fig:exp_1}
\end{figure}

\cref{tab:experimental} shows that our approach outperforms all versions of the baseline. By comparing the \textit{F1 scores} and \textit{Hand Collision Percentage} between DiffGrasp and OMOMO, we can observe that our work can model and ensure more continuous and precise grasping. 
By comparing with OMOMO-V2, we see that jointly modeling the hand pose and body pose based on the baseline yields less dynamic mean poses. And qualitative results in our supplementary materials show that compared with modeling hand pose failure simultaneously leads to a degradation in its ability to model body pose.
Although OMOMO-V3 achieves better results in terms of fine and continuous grasping compared to OMOMO and OMOMO-V2
, it still performs worse than DiffGrasp.
OMOMO and OMOMO-V3 have smaller FS because the generated character body pose tends to be static. As shown on the right side of \cref{fig:exp_1}, DiffGrasp achieves a stooping motion while the baseline remains upright.

\begin{table*}[!hbpt]
    \centering
    \fontsize{9}{10}\selectfont
 \begin{tabular}{l|cccccccccc}

\toprule											
Method	&	Hands JPE$\downarrow$	&	 MPJPE$\downarrow$	&	 MPVPE$\downarrow$	&	 FS$\downarrow$	&	 Coll. \%$\downarrow$	&	 C depth $\downarrow$ 	&		 F1$\uparrow$ 	&	 Cont dist$\downarrow$ \\
\midrule											

Full loss w/o Recon 	& 	25.49 	&	13.69 	&	11.07 	&	2.89 	&	\textbf{0.0001} 	&   0.0001 	&	0.5543 	&	 0.07 	\\

\midrule

Full loss w/o Inter and Recon	&	25.80 	&	14.13 	&	11.42 	&	3.43 	&	0.0022 	&	0.0003 	&	0.3861 	&	0.10\\

Full loss w/o Inter w/ Simp Recon 	& 	23.42 	&	13.32 	&	10.92 	&	\textbf{2.17} 	&	0.0004 	&	0.0001 	&	0.3961 	&	0.07\\

Full loss w/o Inter 	& 	22.35 	&	13.12 	&	10.86 	&	2.20	&	0.0002 	&	0.0001 	&	0.5319 	&	0.06 \\

\midrule

Full loss w/o $PE^P$	& 	21.97 	&	12.66 	&	10.40 	&	2.51 	&	0.0007 	&	0.0001 	&	0.6448 	&	0.05 	\\	

Full loss	& 	21.42 	&	12.48 	&	10.26 	&	2.59 	&	0.0033 	&	0.0002 	&	0.6982 	&	0.05 	\\	

\midrule

\noindent \textbf{DiffGrasp} (Full loss w/ Guidance)	& 	\textbf{20.99} 	&	\textbf{12.24} 	&	\textbf{10.09} 	&	2.22 	&	0.0023 	&	\textbf{0.0001} 	&	\textbf{0.7840} 	&	\textbf{0.04} 	\\	

\bottomrule			

    \end{tabular}

    \caption{\textit{Ablation study results} on GRAB \cite{GRAB:2020} dataset. }
    \label{tab:ablation}
\end{table*}

\subsection{Ablation Study}
We conduct ablation studies on contact-aware losses, frame-wise positional encoding $PE^P$ and our guidance strategy. The experimental results are shown in \cref{tab:ablation} and \cref{fig:ablation}.

\subsubsection{Effect of Contact-aware Losses.}

\begin{figure}[hbtp]
    \centering
    \includegraphics[width=\columnwidth]{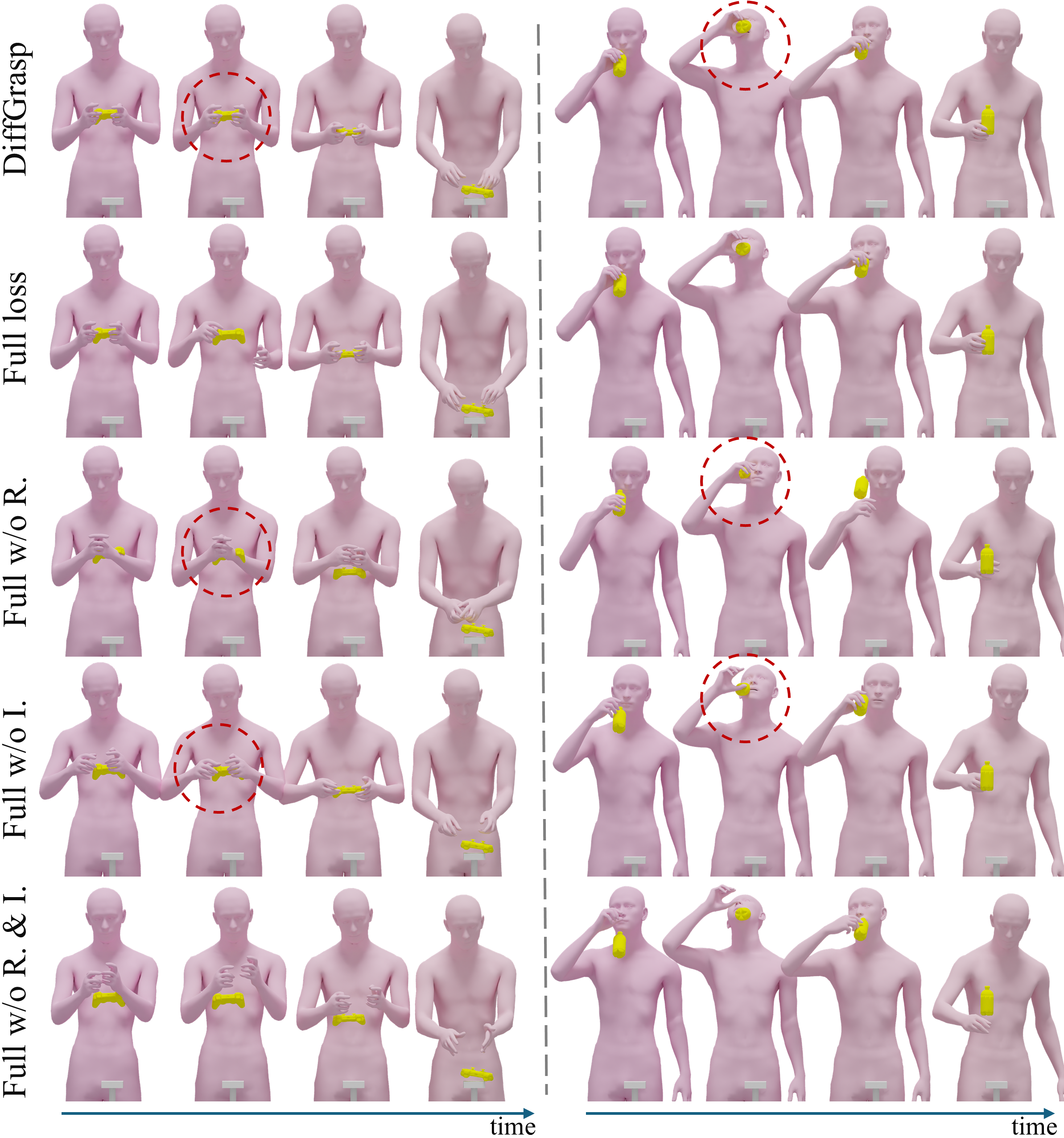}
    \caption{\textit{Qualitative Results of Ablation Study.} 
    In this figure, Full is the abbreviation for Full loss, R. is the abbreviation for Recon, and I. is the abbreviation for Inter.} 
    \label{fig:ablation}
\end{figure}

Ablation studies on \textit{the contact-aware reconstruction loss} show that starting from a version with only the diffusion loss (Full loss w/o Inter and Recon), progressively adding the simple reconstruction loss (Full loss w/o Inter w/ Simp Recon), which does not use the contact label constraint, and then using our contact-aware reconstruction loss (Full loss w/o Inter) results in significant improvements across all metrics.
In the ablation of \textit{the contact-aware interaction loss}, the version with interaction loss term (Full loss w/o Recon) outperforms all other loss ablations in the F1 metric, demonstrating that the interaction loss encourages contact between the hands and the object. Qualitative results in \cref{fig:ablation} shows that the interaction loss (Full w/o R.) brings the generated hand closer to the object but results in less natural poses, while the reconstruction loss (Full w/o I.) generates more natural poses but with weaker perception of the position of object.

\subsubsection{Effect of Frame-wise Positional Encoding.} 
Removing the frame-wise positional encoding $PE^P$ (Full loss w/o $PE^P$)  from the full loss version (Full loss) results in a slight decrease in almost all metrics, demonstrating that $PE^P$ help the network understand the input conditions better.

\subsubsection{Effect of Guidance.}
Comparing our model without guidance (complete loss) with our full model incorporating guidance (DiffGrasp), we can see slight improvements in accuracy, and user experiment results in the supplementary demonstrate that guidance generated more realistic results.

\subsection{Limitations}
\begin{figure}[!htbp]
    \centering
    \includegraphics[width=\columnwidth]{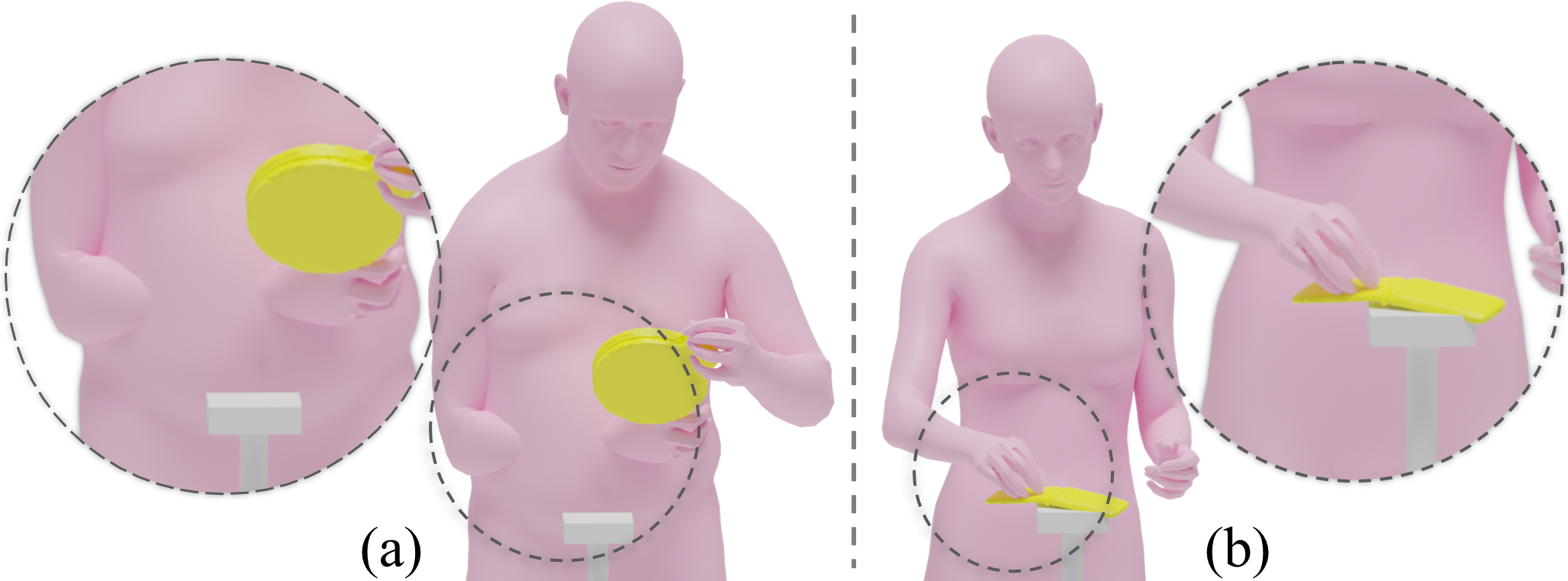}
    \caption{\textit{Limitations.} Our method may generate self-penetration (a) or unrealistic poses (b) in some cases.}
    \label{fig:limitations}
\end{figure}

Although promising results have been achieved, our method has several limitations, as demonstrated in \cref{fig:limitations}.
We find that due to the lack of constraints on the self-penetration of human body meshes, DiffGrasp may generate results with self-penetration (\cref{fig:limitations}(a)). Moreover, because we do not add physical constraints between the hand and object, the generated hand results may exhibit unrealistic grasp poses (\cref{fig:limitations}(b)). 
Furthermore, due to limitations in the dataset used, we have not yet achieved walking during grasping.

\section{Conclusion}

In this paper, we propose the first framework to generate realistic whole-body human motion sequence with fine finger-object grasp, conditioned on a 3D object motion sequence with different object sizes and shapes.
Furthermore, by using reliable contact-aware losses, we leverage a single conditional diffusion model generate natural and reliable results. 
Finally, we utilize an innovative data-driven guidance strategy, along with others, to achieve a stable, non-penetrating, and non-sliding grasp. Extensive experiments show that our method achieves state-of-the-art results.

\section*{Acknowledgments}
This work was supported in part by 
National Science and Technology Major Project (2022ZD0117904), National Natural Science Foundation of China (62473356,62373061), Beijing Natural Science Foundation (L232028), CAS Major Project (RCJJ-145-24-14), and Beijing Hospitals Authority Clinical Medicine Development of Special Funding Support No. ZLRK202330.

\bibliography{aaai25}

\begin{thebibliography}{56}
\providecommand{\natexlab}[1]{#1}

\bibitem[{Ao, Zhang, and Liu(2023)}]{ao2023gesturediffuclip}
Ao, T.; Zhang, Z.; and Liu, L. 2023.
\newblock GestureDiffuCLIP: Gesture diffusion model with CLIP latents.
\newblock \emph{arXiv preprint arXiv:2303.14613}.

\bibitem[{Braun et~al.(2024)Braun, Christen, Kocabas, Aksan, and Hilliges}]{braun2023physically}
Braun, J.; Christen, S.; Kocabas, M.; Aksan, E.; and Hilliges, O. 2024.
\newblock Physically Plausible Full-Body Hand-Object Interaction Synthesis.
\newblock In \emph{International Conference on 3D Vision (3DV)}.

\bibitem[{Cen et~al.(2024)Cen, Pi, Peng, Shen, Yang, Shuai, Bao, and Zhou}]{cen2024text_scene_motion}
Cen, Z.; Pi, H.; Peng, S.; Shen, Z.; Yang, M.; Shuai, Z.; Bao, H.; and Zhou, X. 2024.
\newblock Generating Human Motion in 3D Scenes from Text Descriptions.
\newblock In \emph{CVPR}.

\bibitem[{Cha et~al.(2024)Cha, Kim, Yoon, and Baek}]{cha2024text2hoi}
Cha, J.; Kim, J.; Yoon, J.~S.; and Baek, S. 2024.
\newblock Text2HOI: Text-guided 3D Motion Generation for Hand-Object Interaction.
\newblock In \emph{Proceedings of the IEEE/CVF Conference on Computer Vision and Pattern Recognition}, 1577--1585.

\bibitem[{Chen et~al.(2023)Chen, Jiang, Liu, Huang, Fu, Chen, and Yu}]{chen2023executing}
Chen, X.; Jiang, B.; Liu, W.; Huang, Z.; Fu, B.; Chen, T.; and Yu, G. 2023.
\newblock Executing your Commands via Motion Diffusion in Latent Space.
\newblock In \emph{Proceedings of the IEEE/CVF Conference on Computer Vision and Pattern Recognition}, 18000--18010.

\bibitem[{Christen et~al.(2022)Christen, Kocabas, Aksan, Hwangbo, Song, and Hilliges}]{christen2022dgrasp}
Christen, S.; Kocabas, M.; Aksan, E.; Hwangbo, J.; Song, J.; and Hilliges, O. 2022.
\newblock D-Grasp: Physically Plausible Dynamic Grasp Synthesis for Hand-Object Interactions.
\newblock arXiv:2112.03028.

\bibitem[{Diller and Dai(2024)}]{diller2024cg}
Diller, C.; and Dai, A. 2024.
\newblock Cg-hoi: Contact-guided 3d human-object interaction generation.
\newblock In \emph{Proceedings of the IEEE/CVF Conference on Computer Vision and Pattern Recognition}, 19888--19901.

\bibitem[{Fan et~al.(2023)Fan, Taheri, Tzionas, Kocabas, Kaufmann, Black, and Hilliges}]{fan2023arctic}
Fan, Z.; Taheri, O.; Tzionas, D.; Kocabas, M.; Kaufmann, M.; Black, M.~J.; and Hilliges, O. 2023.
\newblock {ARCTIC}: A Dataset for Dexterous Bimanual Hand-Object Manipulation.
\newblock In \emph{Proceedings IEEE Conference on Computer Vision and Pattern Recognition (CVPR)}.

\bibitem[{Ghosh et~al.(2023{\natexlab{a}})Ghosh, Dabral, Golyanik, Theobalt, and Slusallek}]{ghosh2022imos}
Ghosh, A.; Dabral, R.; Golyanik, V.; Theobalt, C.; and Slusallek, P. 2023{\natexlab{a}}.
\newblock IMoS: Intent-Driven Full-Body Motion Synthesis for Human-Object Interactions.
\newblock In \emph{Eurographics}.

\bibitem[{Ghosh et~al.(2023{\natexlab{b}})Ghosh, Dabral, Golyanik, Theobalt, and Slusallek}]{ghosh2023remos}
Ghosh, A.; Dabral, R.; Golyanik, V.; Theobalt, C.; and Slusallek, P. 2023{\natexlab{b}}.
\newblock ReMoS: Reactive 3D Motion Synthesis for Two-Person Interactions.
\newblock arXiv:2311.17057.

\bibitem[{Guo et~al.(2022)Guo, Zuo, Wang, and Cheng}]{guo2022tm2t}
Guo, C.; Zuo, X.; Wang, S.; and Cheng, L. 2022.
\newblock Tm2t: Stochastic and tokenized modeling for the reciprocal generation of 3d human motions and texts.
\newblock In \emph{European Conference on Computer Vision}, 580--597. Springer.

\bibitem[{Hasson et~al.(2019)Hasson, Varol, Tzionas, Kalevatykh, Black, Laptev, and Schmid}]{hasson2019learning}
Hasson, Y.; Varol, G.; Tzionas, D.; Kalevatykh, I.; Black, M.~J.; Laptev, I.; and Schmid, C. 2019.
\newblock Learning joint reconstruction of hands and manipulated objects.
\newblock arXiv:1904.05767.

\bibitem[{He et~al.(2022)He, Saito, Zachary, Rushmeier, and Zhou}]{he2022nemf}
He, C.; Saito, J.; Zachary, J.; Rushmeier, H.; and Zhou, Y. 2022.
\newblock Nemf: Neural motion fields for kinematic animation.
\newblock \emph{Advances in Neural Information Processing Systems}, 35: 4244--4256.

\bibitem[{Ho, Jain, and Abbeel(2020)}]{ho2020denoising}
Ho, J.; Jain, A.; and Abbeel, P. 2020.
\newblock Denoising Diffusion Probabilistic Models.
\newblock arXiv:2006.11239.

\bibitem[{Ho et~al.(2022)Ho, Salimans, Gritsenko, Chan, Norouzi, and Fleet}]{ho2022video}
Ho, J.; Salimans, T.; Gritsenko, A.; Chan, W.; Norouzi, M.; and Fleet, D.~J. 2022.
\newblock Video diffusion models.
\newblock \emph{arXiv:2204.03458}.

\bibitem[{Huang et~al.(2023)Huang, Wang, Li, Jia, Liu, Zhu, Liang, and Zhu}]{huang2023diffusion}
Huang, S.; Wang, Z.; Li, P.; Jia, B.; Liu, T.; Zhu, Y.; Liang, W.; and Zhu, S.-C. 2023.
\newblock Diffusion-based Generation, Optimization, and Planning in 3D Scenes.
\newblock In \emph{Proceedings of the IEEE/CVF Conference on Computer Vision and Pattern Recognition (CVPR)}.

\bibitem[{Jiang et~al.(2021)Jiang, Liu, Wang, and Wang}]{jiang2021handobject}
Jiang, H.; Liu, S.; Wang, J.; and Wang, X. 2021.
\newblock Hand-Object Contact Consistency Reasoning for Human Grasps Generation.
\newblock arXiv:2104.03304.

\bibitem[{Kulkarni et~al.(2023)Kulkarni, Rempe, Genova, Kundu, Johnson, Fouhey, and Guibas}]{kulkarni2023nifty}
Kulkarni, N.; Rempe, D.; Genova, K.; Kundu, A.; Johnson, J.; Fouhey, D.; and Guibas, L. 2023.
\newblock NIFTY: Neural Object Interaction Fields for Guided Human Motion Synthesis.
\newblock arXiv:2307.07511.

\bibitem[{Kwon et~al.(2021)Kwon, Tekin, St\"uhmer, Bogo, and Pollefeys}]{H2o_2021_ICCV}
Kwon, T.; Tekin, B.; St\"uhmer, J.; Bogo, F.; and Pollefeys, M. 2021.
\newblock H2O: Two Hands Manipulating Objects for First Person Interaction Recognition.
\newblock In \emph{Proceedings of the IEEE/CVF International Conference on Computer Vision (ICCV)}, 10138--10148.

\bibitem[{Lee, Moon, and Lee(2023)}]{lee2023multiact}
Lee, T.; Moon, G.; and Lee, K.~M. 2023.
\newblock MultiAct: Long-term 3D human motion generation from multiple action labels.
\newblock In \emph{Proceedings of the AAAI Conference on Artificial Intelligence}, volume 37-1, 1231--1239.

\bibitem[{Li et~al.(2023{\natexlab{a}})Li, Lin, Zhou, Li, Huo, Chen, and Ye}]{li2023contact2grasp}
Li, H.; Lin, X.; Zhou, Y.; Li, X.; Huo, Y.; Chen, J.; and Ye, Q. 2023{\natexlab{a}}.
\newblock Contact2Grasp: 3D Grasp Synthesis via Hand-Object Contact Constraint.
\newblock arXiv:2210.09245.

\bibitem[{Li et~al.(2023{\natexlab{b}})Li, Clegg, Mottaghi, Wu, Puig, and Liu}]{li2023controllable}
Li, J.; Clegg, A.; Mottaghi, R.; Wu, J.; Puig, X.; and Liu, C.~K. 2023{\natexlab{b}}.
\newblock Controllable human-object interaction synthesis.
\newblock \emph{arXiv preprint arXiv:2312.03913}.

\bibitem[{Li, Wu, and Liu(2023)}]{li2023object}
Li, J.; Wu, J.; and Liu, C.~K. 2023.
\newblock Object motion guided human motion synthesis.
\newblock \emph{ACM Transactions on Graphics (TOG)}, 42(6): 1--11.

\bibitem[{Li et~al.(2024)Li, Wang, Loy, and Dai}]{Li_2024_WACV}
Li, Q.; Wang, J.; Loy, C.~C.; and Dai, B. 2024.
\newblock Task-Oriented Human-Object Interactions Generation With Implicit Neural Representations.
\newblock In \emph{Proceedings of the IEEE/CVF Winter Conference on Applications of Computer Vision (WACV)}, 3035--3044.

\bibitem[{Li et~al.(2023{\natexlab{c}})Li, Zhao, Zhang, Su, Ren, Zhang, Tang, and Li}]{li2023finedance}
Li, R.; Zhao, J.; Zhang, Y.; Su, M.; Ren, Z.; Zhang, H.; Tang, Y.; and Li, X. 2023{\natexlab{c}}.
\newblock FineDance: A Fine-grained Choreography Dataset for 3D Full Body Dance Generation.
\newblock In \emph{Proceedings of the IEEE/CVF International Conference on Computer Vision}, 10234--10243.

\bibitem[{Liu et~al.(2023)Liu, Zhou, Yang, Gupta, and Wang}]{liu2023contactgen}
Liu, S.; Zhou, Y.; Yang, J.; Gupta, S.; and Wang, S. 2023.
\newblock ContactGen: Generative Contact Modeling for Grasp Generation.
\newblock In \emph{Proceedings of the IEEE/CVF International Conference on Computer Vision}.

\bibitem[{Liu and Yi(2024)}]{liu2024geneoh}
Liu, X.; and Yi, L. 2024.
\newblock GeneOH Diffusion: Towards Generalizable Hand-Object Interaction Denoising via Denoising Diffusion.
\newblock \emph{arXiv preprint arXiv:2402.14810}.

\bibitem[{Liu et~al.(2024)Liu, Cao, Wen, Jiang, and Ding}]{liu2024towards}
Liu, Y.; Cao, Q.; Wen, Y.; Jiang, H.; and Ding, C. 2024.
\newblock Towards Variable and Coordinated Holistic Co-Speech Motion Generation.
\newblock In \emph{Proceedings of the IEEE/CVF Conference on Computer Vision and Pattern Recognition}, 1566--1576.

\bibitem[{Lu et~al.(2023)Lu, Chen, Zeng, Lin, Zhang, Zhang, and Shum}]{lu2023humantomato}
Lu, S.; Chen, L.-H.; Zeng, A.; Lin, J.; Zhang, R.; Zhang, L.; and Shum, H.-Y. 2023.
\newblock Humantomato: Text-aligned whole-body motion generation.
\newblock \emph{arXiv preprint arXiv:2310.12978}.

\bibitem[{Lucas et~al.(2022)Lucas, Baradel, Weinzaepfel, and Rogez}]{lucas2022posegpt}
Lucas, T.; Baradel, F.; Weinzaepfel, P.; and Rogez, G. 2022.
\newblock Posegpt: Quantization-based 3d human motion generation and forecasting.
\newblock In \emph{European Conference on Computer Vision}, 417--435. Springer.

\bibitem[{Luo, Liu, and Yi(2024)}]{luo2024physics}
Luo, H.; Liu, Y.; and Yi, L. 2024.
\newblock Physics-aware Hand-object Interaction Denoising.
\newblock In \emph{Proceedings of the IEEE/CVF Conference on Computer Vision and Pattern Recognition}, 2341--2350.

\bibitem[{Pavlakos et~al.(2019)Pavlakos, Choutas, Ghorbani, Bolkart, Osman, Tzionas, and Black}]{SMPL-X:2019}
Pavlakos, G.; Choutas, V.; Ghorbani, N.; Bolkart, T.; Osman, A. A.~A.; Tzionas, D.; and Black, M.~J. 2019.
\newblock Expressive Body Capture: 3D Hands, Face, and Body from a Single Image.
\newblock In \emph{Proceedings IEEE Conf. on Computer Vision and Pattern Recognition (CVPR)}.

\bibitem[{Peng et~al.(2023)Peng, Xie, Wu, Jampani, Sun, and Jiang}]{peng2023hoi}
Peng, X.; Xie, Y.; Wu, Z.; Jampani, V.; Sun, D.; and Jiang, H. 2023.
\newblock HOI-Diff: Text-Driven Synthesis of 3D Human-Object Interactions using Diffusion Models.
\newblock \emph{arXiv preprint arXiv:2312.06553}.

\bibitem[{Petrovich, Black, and Varol(2022)}]{petrovich2022temos}
Petrovich, M.; Black, M.~J.; and Varol, G. 2022.
\newblock TEMOS: Generating diverse human motions from textual descriptions.
\newblock In \emph{European Conference on Computer Vision}, 480--497. Springer.

\bibitem[{Pinyoanuntapong et~al.(2024)Pinyoanuntapong, Wang, Lee, and Chen}]{pinyoanuntapong2024mmm}
Pinyoanuntapong, E.; Wang, P.; Lee, M.; and Chen, C. 2024.
\newblock Mmm: Generative masked motion model.
\newblock In \emph{Proceedings of the IEEE/CVF Conference on Computer Vision and Pattern Recognition}, 1546--1555.

\bibitem[{Prokudin, Lassner, and Romero(2019)}]{prokudin2019efficient}
Prokudin, S.; Lassner, C.; and Romero, J. 2019.
\newblock Efficient Learning on Point Clouds With Basis Point Sets.
\newblock In \emph{Proceedings of the IEEE International Conference on Computer Vision}, 4332--4341.

\bibitem[{Sahbani, El-Khoury, and Bidaud(2012)}]{sahbani2012overview}
Sahbani, A.; El-Khoury, S.; and Bidaud, P. 2012.
\newblock An overview of 3D object grasp synthesis algorithms.
\newblock \emph{Robotics and Autonomous Systems}, 60(3): 326--336.

\bibitem[{Shafir et~al.(2023)Shafir, Tevet, Kapon, and Bermano}]{shafir2023human}
Shafir, Y.; Tevet, G.; Kapon, R.; and Bermano, A.~H. 2023.
\newblock Human Motion Diffusion as a Generative Prior.
\newblock arXiv:2303.01418.

\bibitem[{Song et~al.(2024)Song, Zhang, Li, Gao, Hao, Hou, Chen, Li, and Qin}]{song2024hoianimator}
Song, W.; Zhang, X.; Li, S.; Gao, Y.; Hao, A.; Hou, X.; Chen, C.; Li, N.; and Qin, H. 2024.
\newblock HOIAnimator: Generating Text-prompt Human-object Animations using Novel Perceptive Diffusion Models.
\newblock In \emph{Proceedings of the IEEE/CVF Conference on Computer Vision and Pattern Recognition}, 811--820.

\bibitem[{Taheri et~al.(2022)Taheri, Choutas, Black, and Tzionas}]{taheri2022goal}
Taheri, O.; Choutas, V.; Black, M.~J.; and Tzionas, D. 2022.
\newblock GOAL: Generating 4D whole-body motion for hand-object grasping.
\newblock In \emph{Proceedings of the IEEE/CVF Conference on Computer Vision and Pattern Recognition}, 13263--13273.

\bibitem[{Taheri et~al.(2020)Taheri, Ghorbani, Black, and Tzionas}]{GRAB:2020}
Taheri, O.; Ghorbani, N.; Black, M.~J.; and Tzionas, D. 2020.
\newblock {GRAB}: A Dataset of Whole-Body Human Grasping of Objects.
\newblock In \emph{European Conference on Computer Vision (ECCV)}.

\bibitem[{Taheri et~al.(2023)Taheri, Zhou, Tzionas, Zhou, Ceylan, Pirk, and Black}]{taheri2023grip}
Taheri, O.; Zhou, Y.; Tzionas, D.; Zhou, Y.; Ceylan, D.; Pirk, S.; and Black, M.~J. 2023.
\newblock Grip: Generating interaction poses using latent consistency and spatial cues.
\newblock \emph{arXiv preprint arXiv:2308.11617}.

\bibitem[{Tendulkar, Sur{\'\i}s, and Vondrick(2023)}]{tendulkar2023flex}
Tendulkar, P.; Sur{\'\i}s, D.; and Vondrick, C. 2023.
\newblock Flex: Full-body grasping without full-body grasps.
\newblock In \emph{Proceedings of the IEEE/CVF Conference on Computer Vision and Pattern Recognition}, 21179--21189.

\bibitem[{Tevet et~al.(2023)Tevet, Raab, Gordon, Shafir, Cohen-or, and Bermano}]{tevet2023human}
Tevet, G.; Raab, S.; Gordon, B.; Shafir, Y.; Cohen-or, D.; and Bermano, A.~H. 2023.
\newblock Human Motion Diffusion Model.
\newblock In \emph{The Eleventh International Conference on Learning Representations}.

\bibitem[{Tseng, Castellon, and Liu(2023)}]{tseng2023edge}
Tseng, J.; Castellon, R.; and Liu, K. 2023.
\newblock Edge: Editable dance generation from music.
\newblock In \emph{Proceedings of the IEEE/CVF Conference on Computer Vision and Pattern Recognition}, 448--458.

\bibitem[{Vaswani et~al.(2017)Vaswani, Shazeer, Parmar, Uszkoreit, Jones, Gomez, Kaiser, and Polosukhin}]{vaswani2017attention}
Vaswani, A.; Shazeer, N.; Parmar, N.; Uszkoreit, J.; Jones, L.; Gomez, A.~N.; Kaiser, {\L}.; and Polosukhin, I. 2017.
\newblock Attention is all you need.
\newblock \emph{Advances in neural information processing systems}, 30.

\bibitem[{Wu et~al.(2022)Wu, Wang, Zhang, Zhang, Hilliges, Yu, and Tang}]{wu2022saga}
Wu, Y.; Wang, J.; Zhang, Y.; Zhang, S.; Hilliges, O.; Yu, F.; and Tang, S. 2022.
\newblock SAGA: Stochastic Whole-Body Grasping with Contact.
\newblock In \emph{Proceedings of the European Conference on Computer Vision (ECCV)}.

\bibitem[{Xu et~al.(2023)Xu, Li, Wang, and Gui}]{interdiff}
Xu, S.; Li, Z.; Wang, Y.-X.; and Gui, L.-Y. 2023.
\newblock {InterDiff}: Generating 3D Human-Object Interactions with Physics-Informed Diffusion.
\newblock In \emph{ICCV}.

\bibitem[{Zhang et~al.(2024{\natexlab{a}})Zhang, Christen, Fan, Zheng, Hwangbo, Song, and Hilliges}]{zhang2024artigrasp}
Zhang, H.; Christen, S.; Fan, Z.; Zheng, L.; Hwangbo, J.; Song, J.; and Hilliges, O. 2024{\natexlab{a}}.
\newblock {ArtiGrasp}: Physically Plausible Synthesis of Bi-Manual Dexterous Grasping and Articulation.
\newblock In \emph{International Conference on 3D Vision (3DV)}.

\bibitem[{Zhang et~al.(2021)Zhang, Ye, Shiratori, and Komura}]{zhang2021manipnet}
Zhang, H.; Ye, Y.; Shiratori, T.; and Komura, T. 2021.
\newblock Manipnet: neural manipulation synthesis with a hand-object spatial representation.
\newblock \emph{ACM Transactions on Graphics (ToG)}, 40(4): 1--14.

\bibitem[{Zhang et~al.(2024{\natexlab{b}})Zhang, Cai, Pan, Hong, Guo, Yang, and Liu}]{zhang2024motiondiffuse}
Zhang, M.; Cai, Z.; Pan, L.; Hong, F.; Guo, X.; Yang, L.; and Liu, Z. 2024{\natexlab{b}}.
\newblock Motiondiffuse: Text-driven human motion generation with diffusion model.
\newblock \emph{IEEE Transactions on Pattern Analysis and Machine Intelligence}.

\bibitem[{Zhao et~al.(2023)Zhao, Zhang, Wang, Beeler, and Tang}]{zhao2023synthesizing}
Zhao, K.; Zhang, Y.; Wang, S.; Beeler, T.; and Tang, S. 2023.
\newblock Synthesizing diverse human motions in 3d indoor scenes.
\newblock In \emph{Proceedings of the IEEE/CVF International Conference on Computer Vision}, 14738--14749.

\bibitem[{Zheng et~al.(2023{\natexlab{a}})Zheng, Zheng, Fang, Liu, and Yi}]{zheng2023cams}
Zheng, J.; Zheng, Q.; Fang, L.; Liu, Y.; and Yi, L. 2023{\natexlab{a}}.
\newblock Cams: Canonicalized manipulation spaces for category-level functional hand-object manipulation synthesis.
\newblock In \emph{Proceedings of the IEEE/CVF Conference on Computer Vision and Pattern Recognition}, 585--594.

\bibitem[{Zheng et~al.(2023{\natexlab{b}})Zheng, Shi, Cui, Zhao, Luo, and Zhou}]{COOP}
Zheng, Y.; Shi, Y.; Cui, Y.; Zhao, Z.; Luo, Z.; and Zhou, W. 2023{\natexlab{b}}.
\newblock COOP: Decoupling and Coupling of Whole-Body Grasping Pose Generation.
\newblock In \emph{Proceedings of the IEEE/CVF International Conference on Computer Vision (ICCV)}, 2163--2173.

\bibitem[{Zhou et~al.(2022)Zhou, Bhatnagar, Lenssen, and Pons-Moll}]{zhou2022toch}
Zhou, K.; Bhatnagar, B.~L.; Lenssen, J.~E.; and Pons-Moll, G. 2022.
\newblock TOCH: Spatio-Temporal Object Correspondence to Hand for Motion Refinement.
\newblock In \emph{European Conference on Computer Vision ({ECCV})}. {Springer}.

\bibitem[{Zhou et~al.(2020)Zhou, Barnes, Lu, Yang, and Li}]{zhou2020continuity}
Zhou, Y.; Barnes, C.; Lu, J.; Yang, J.; and Li, H. 2020.
\newblock On the Continuity of Rotation Representations in Neural Networks.
\newblock arXiv:1812.07035.

\end{thebibliography}

\clearpage

\renewcommand{\thefigure}{A\arabic{figure}}  
\renewcommand{\theequation}{A\arabic{equation}} 
\renewcommand{\thetable}{A\arabic{table}} 
\setcounter{secnumdepth}{2}

\section*{Appendix}
\appendix

In this supplementary, we provide additional qualitative and quantitative analysis for the article, along with more information for reproducibility. We include data preparation for our experiments (\cref{sec:data prepatation}), analyze and compare more potential related work (\cref{sec:More Related Work}), present our implementation details (\cref{sec: impementation details}), and the implementation details for baselines (\cref{sec:implementation for baselines}). We also provide results of human perceptual studies from our method and comparative work (\cref{sec: human perceptual study}), results with unseen objects (\cref{sec:results with unseen}), and more qualitative results and analysis (\cref{sec:more results}). Finally, we introduce a simplified input condition (\cref{sec:simp condition}). 
For further results, please refer to the accompanying supplementary video.

\section{Data Preparation}
\label{sec:data prepatation}
For the GRAB \cite{GRAB:2020} and ARCTIC datasets \cite{fan2023arctic}, we partition the data by subjects: s1 to s8 are used for training, while s9 and s10 are used for testing. In the GRAB training dataset, we removed 5 objects to conduct experiments with unseen objects. For both datasets, we downsample training data to 15 Hz, so the sequence duration is 2s. We used a sliding-window approach to segment each interaction sequence in the dataset and trimmed the sequences to remove frames where the interaction had ended or had not yet started. Each segmented training sequence was normalized by aligning its first frame’s human body x and y coordinates to the origin. We then calculated the mean and standard deviation of all sequences in the dataset for normalization during training. 

Finally, the GRAB dataset contains 28,440 training and 3,378 testing sequences, while the ARCTIC dataset contains 33,430 training and 3,168 testing sequences.

\section{More Related Work}
\label{sec:More Related Work}
In this section, we will introduce three different types of related work, all of which produce results very similar to those generated by DiffGrasp. First, with respect to the intent-driven method, IMoS \cite{ghosh2022imos}, we made modifications to its network to adapt it to our task, but unfortunately it failed to converge and is therefore not included in the main comparison. Second, concerning the work on generating single frame whole-body grasping, we use COOP \cite{COOP} as an example to illustrate its lack of temporal continuity modeling. Finally, regarding recent physics-based generation work \cite{braun2023physically}, while it achieves physically realistic grasping, the generated results are still not sufficiently coherent and fail to model the interaction between both hands.

\label{sec:more related work}

\subsection{Intent-driven Work: IMoS}

We aim to highlight the differences between our method and the previous intent-driven approach. We select IMoS \cite{ghosh2022imos}, which synthesizes full-body pose sequences and 3D object positions from textual 'intent' inputs in an auto-regressive fashion based on two VAE networks. 

We conduct comparison experiments with IMoS, implement with the author’s official code by replacing the input conditions with ours and setting their sequence length $T$ to 30 (as our settings). Unfortunately, the network does not converge on the validation set under our conditional input and data split, so we cannot perform testing and comparison with it. We also observe that IMoS generates fine human poses at the beginning of the generated motion sequence but may fail at the end.
In the following \cref{sec:implementation for baselines}, we provide a detailed explanation of our modification for the IMoS network and its model size.

We believe that DiffGrasp and IMoS are fundamentally suited to different task requirements, making it difficult to adapt the backbone of IMoS by simply modifying the condition inputs as DiffGrasp (Ours). Our task setting is crucial for animation and humanoid robotics, e.g. human motion planning given object keyframe motions. 
In contrast, IMoS aims to generate diverse human and object motion driven by interaction intent. Moreover, IMoS can not achieve the users’ desired object motion.

\begin{figure}[!htbp]
    \centering
    \includegraphics[width=\columnwidth]{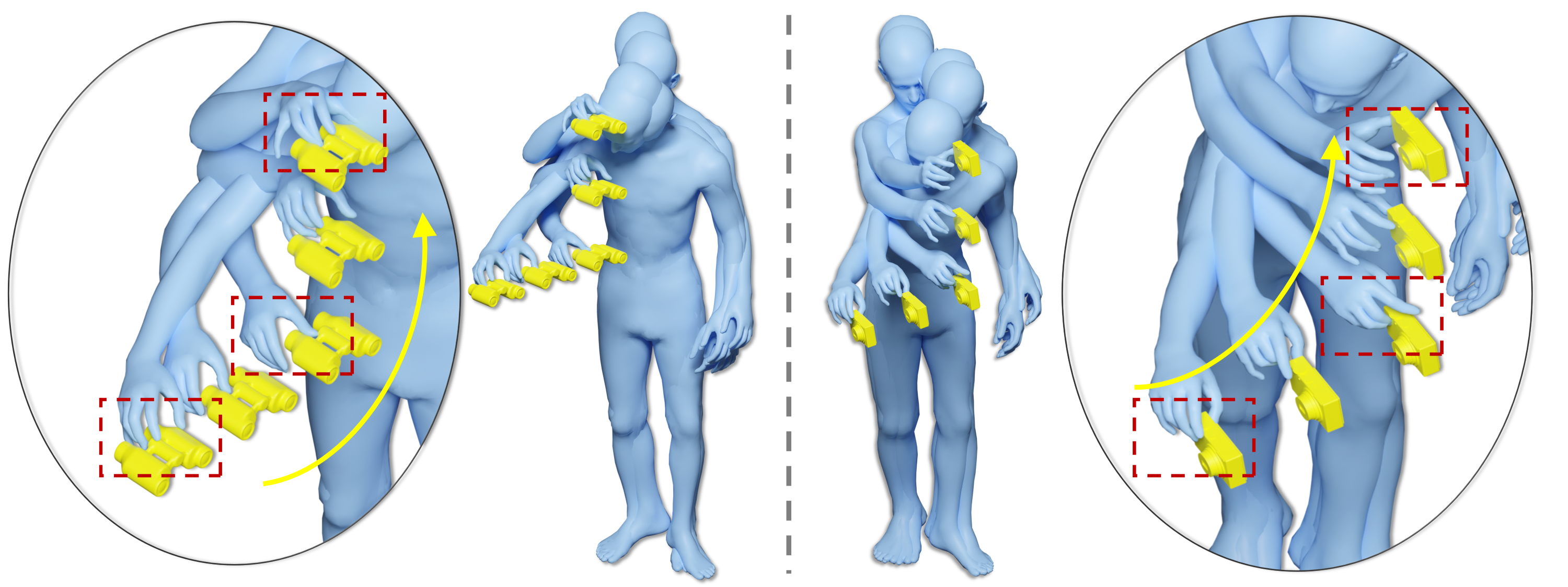}
    \caption{\textit{Qualitative results of single-frame generation method. }The results lack temporal continuity. The red dashed box highlights unreal and discontinuous grasp results of COOP.}
    \label{fig:coop}
\end{figure}

\subsection{Single Frame Generation Work: COOP }
Another task setting similar to ours is single frame generation work, such as SAGA \cite{wu2022saga}, \cite{taheri2022goal}, and COOP \cite{COOP}. 
These works can generate natural single frame whole-body grasping results based on the 3D positions of different small objects. However, they cannot perform well in our setting due to their lack of temporal modeling capabilities. 
In order to further evaluate the temporal generation ability, we employed the state-of-the-art single-frame work COOP \cite{COOP} running on each object motion frame. The results are shown in \cref{fig:coop}. Although the grasping outcomes are fine, they evidently lack temporal continuity.

\subsection{Physics-based Generation Work }
Recently, physically plausible full-body grasping synthesis is proposed \cite{braun2023physically} to generate realistic human-object interaction sequences. Although effective, this approach cannot generate bi-manual full-body grasping due to the high degrees-of-freedom inherent in bi-manual manipulation.

\section{Implementation Details}
\label{sec: impementation details}
We use $N = 1,000$ noising steps. We use sinusoidal positional encoding for our frame-wise and part-wise positional encodings. We choose a fixed length of motion sequences $T = 30$. For each weight term in the final loss function, we set $\lambda_{diff}=1$, $\lambda_{recon}=2$ and $\lambda_{inter}=1$. For all ablation studies, we use the Adam optimizer with a learning rate of $1e-4$, training for 300 epochs. The training was carried out on two NVIDIA A100 GPUs, which takes about 48 hours to complete.

\subsection{Model Architecture }

\begin{figure*}[!htbp]
    \centering
    \includegraphics[width=\textwidth]{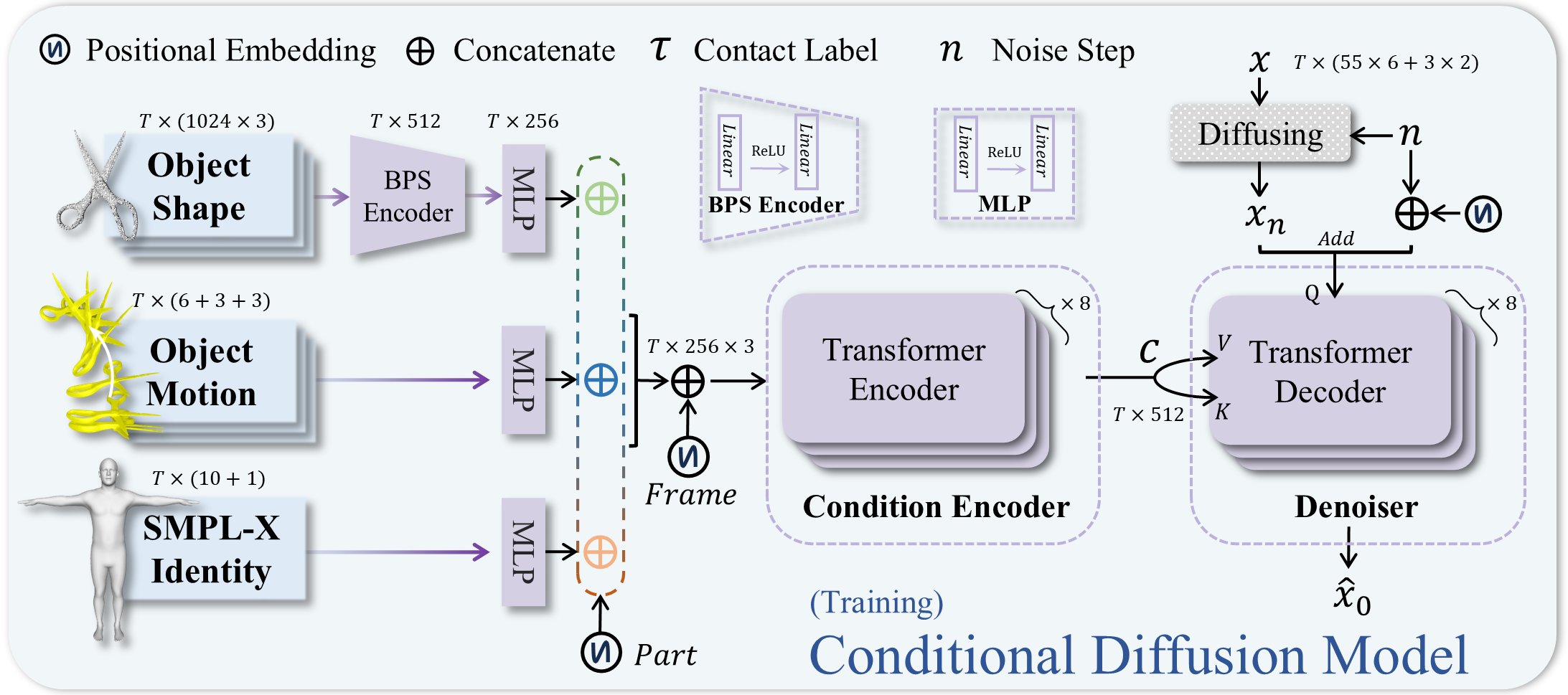}
    \caption{\textit{Model Architecture.} The BPS Encoder and MLP are simple stacks of linear layers for feature alignment, while the condition encoder and denoiser use a standard 8-layer transformer design. The diffusion process (Diffusing module) follows the standard DDPM formula. }
    \label{fig:arch}
\end{figure*}

Our conditional diffusion model is shown in \cref{fig:arch}. Both the BPS Encoder and MLP are simple stacks of linear layers and activation functions for feature alignment. The condition encoder and denoiser utilize an 8-layer stacked standard design of transformer encoder and decoder, performing self-attention and multi-head attention, respectively. 

The diffusion process (the Diffusing module) follows the standard DDPM diffusing formula. During the inference stage, the network receives paired conditions, which are encoded through the MLP and the condition encoder. The encoded condition $c$ and a randomly sampled $x_n$ from a standard normal distribution are then jointly fed into the denoiser to obtain $x_0$. Through a continuous process of noise addition and denoising, the final $x_0$ is generated.

\subsection{Guidance Terms }
For different guidance terms, we use different learning rates and iteration counts to balance the adherence to the model's generated results and realism. For grasp stabilization guidance $\mathcal{G}_{GS}$, we use a learning rate of $1e^{-4}$ with 300 iterations. For hand-object contact guidance $\mathcal{G}_{HO}$, we apply a learning rate of $1e^{-4}$ with 100 iterations. For feet penetration guidance $\mathcal{G}_{Feet}$, we use a learning rate of $1e^{-3}$ with 50 iterations.

Similar to InterDiff \cite{interdiff}, we adopt a strategy that blends single-step generation results $\Tilde{x}$ and the guiding results $\hat{x}$ according to the denoising stage. Given noise step $n$, the optimized result for this step is: 

$$\Tilde{x}_{n-1} = \textit{diff} (   \frac{n}{N}  \times \Tilde{x}_{n} + (1- \frac{n}{N}) \times \hat{x}_{n})$$
where $\textit{diff}()$ is the denoising stage of diffusion.

\section{Implementation Details for Baselines }
\label{sec:implementation for baselines}

\begin{figure*}[!htbp]
    \centering
    \includegraphics[width=\textwidth]{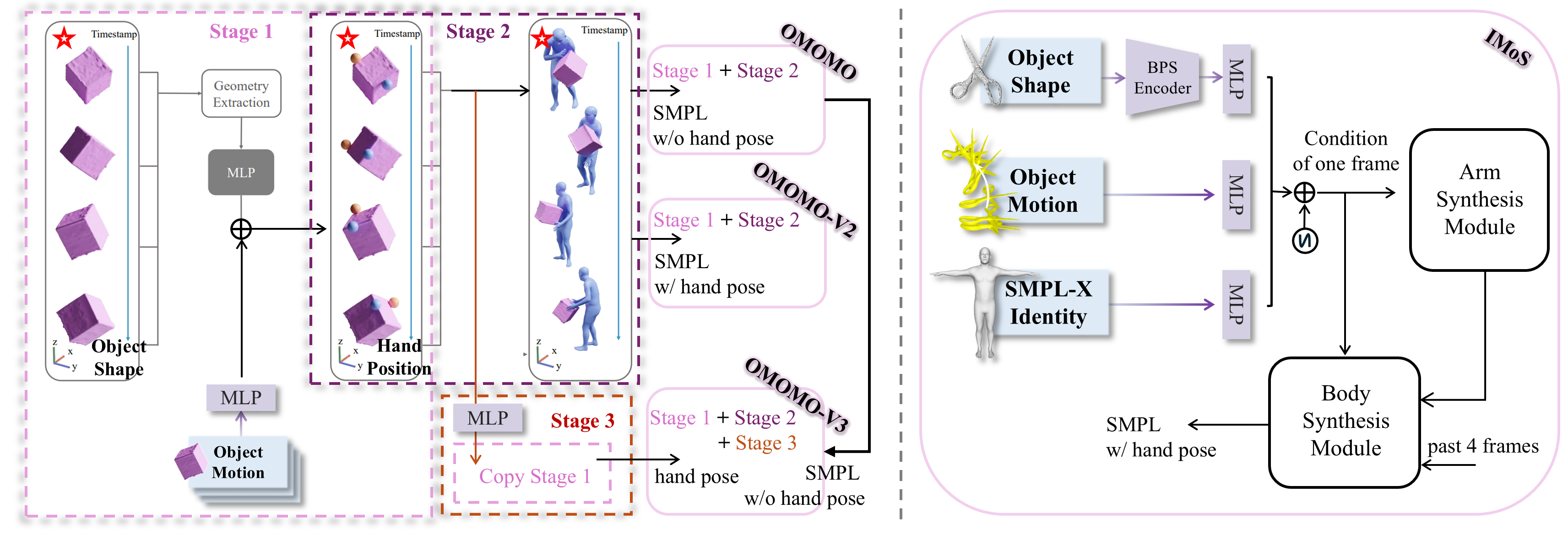}
    \caption{\textit{Implementation Details for Baselines.} On the left, we show our adaptations of the three OMOMO versions. On the right, we show our adaptations of IMoS. The adaptations mainly focus on aligning the network's input conditions to be consistent with the inputs used in our method. }
    \label{fig:details_baseline}
\end{figure*}

\subsubsection{Model Size.}
All models have comparable model sizes: Ours (33.0M), OMOMO (23.5M), OMOMO-V2 (23.9M), OMOMO-V3 (36.2M), IMOS (47.4M).

\subsubsection{Implementation Details for OMOMOs. }
 
On the left side of \cref{fig:details_baseline}, we show the different input-output combinations and the stacking of diffusion models for our three versions of OMOMO. We incorporated the condition embedding layer from our method into OMOMO's input to ensure both methods have the same input. We follow the OMOMO setting, which does not use identity (shape of SMPLX) parameters. Simultaneously, we do not use the \textit{contact 
constraints} post-processing. The post-processing defines any hand within a certain threshold distance from an object as a grasp, which aims to prevent subsequent release in the sequence. However, the training data contains abundant sequences involving releasing and switching hands. The post-processing would negatively affect the generated results of OMOMO.

\subsubsection{Implementation Details for IMoS. }
On the right side of \cref{fig:details_baseline}, we show the adapted version of IMoS.
Since IMoS generates frames autoregressively, meaning it generates the next frame based on the previous four frames, we input a single-frame condition into the IMoS networks, using the same condition encoding network as our method. Following the design of IMoS, we use the ground truth for hand poses, generating the arm pose first and then the body pose. 
In order to maintain consistent with the setting of our task, we did not utilize the \textit{object optimization module} in IMoS to optimizes the position of object.

\section{Human Perceptual Study}
\label{sec: human perceptual study}
We conducted a human perceptual study to further evaluate our method, focusing on evaluating the motion quality and realism of the human-object interaction.   

\begin{figure}[!htbp]
    \centering    \includegraphics[width=\columnwidth]{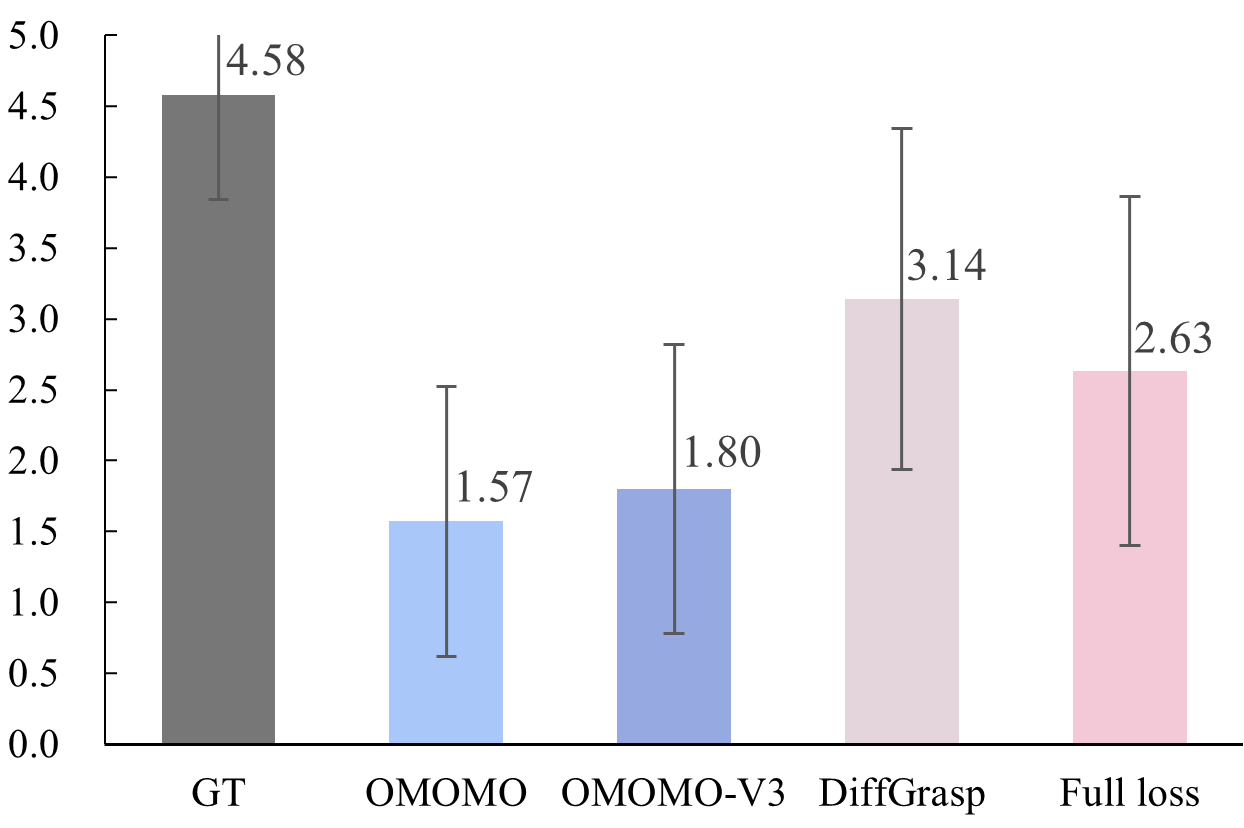}
    \caption{\textit{Results of Human Perceptual Study.} Comparing with OMOMO and our ablation study, DiffGrasp can generate the grasp sequences with the highest score.}
    \label{fig:HumanPerceptualStudy}
\end{figure}

We randomly selected object motion sequences in the test dataset as conditions. For each method, including DiffGrasp (Full loss w/ Guidance), Full loss, OMOMO, OMOMO-V3 along with ground truth,  we generated human motion sequences as videos. 
We put the generated results in a random order and invite participants to score them from 1 to 5 based on the naturalness and realism of the interaction. 
Finally, we selected 20 object motion sequences and invited 20 participants to rate these 5 results and got 2000 ($20\times20\times5$) scores in total. We analyze the scores of each method and the final result is shown in \cref{fig:HumanPerceptualStudy}.
We can observe that our method achieved a higher score than OMOMOs. 

\section{Results with Unseen Objects}
\label{sec:results with unseen}

\begin{figure*}[!hbpt]
    \centering
    \includegraphics[width=0.9\textwidth]{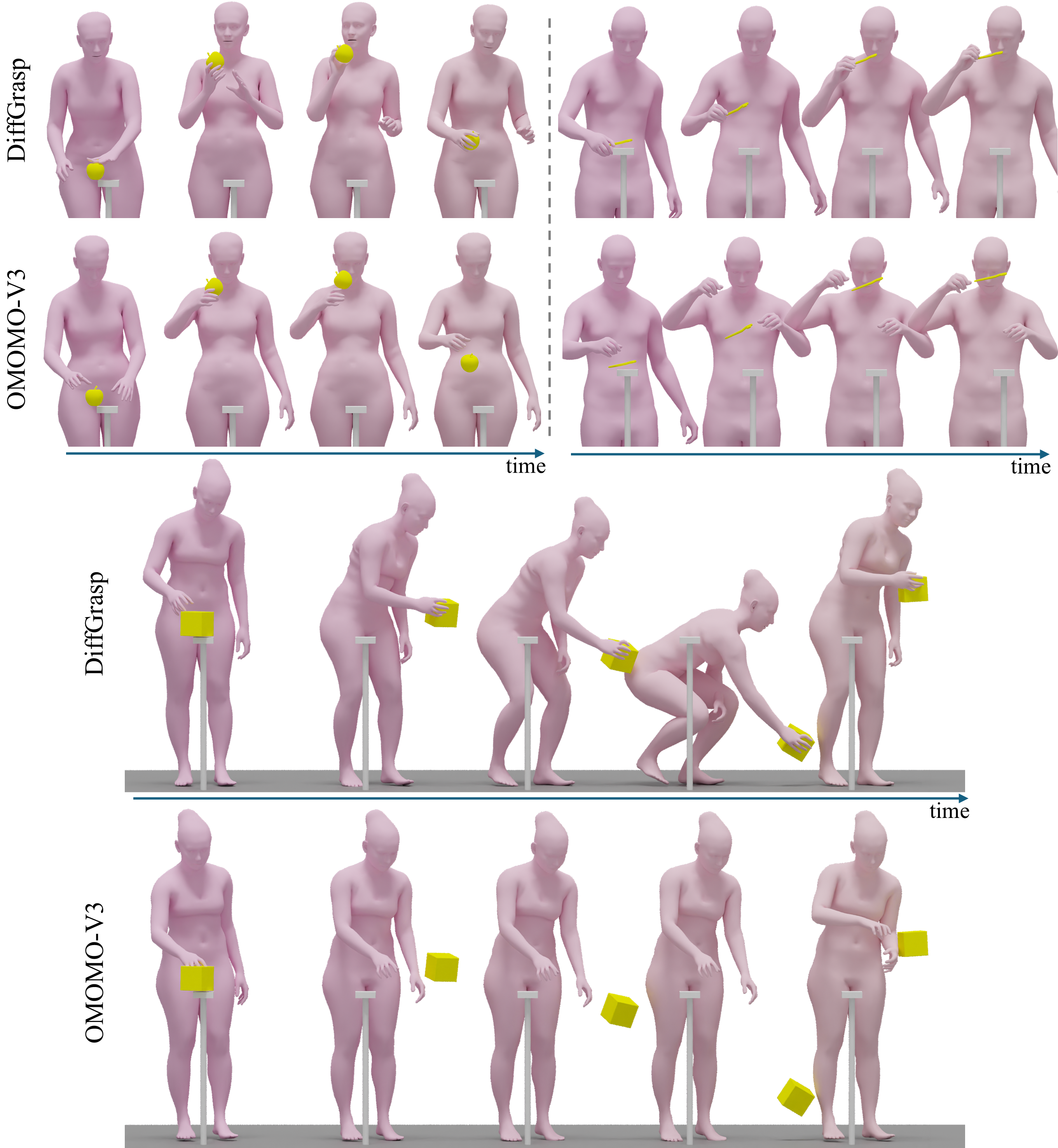}
    \caption{\textit{Qualitative results of unseen objects.} Our method (DiffGrasp) demonstrates strong generalization in grasping unseen objects of various sizes and shapes, while also exhibiting excellent full-body coordination. }
    \label{fig:unknownobj}
\end{figure*}

To further evaluate our model’s generalization ability to unseen objects, we exclude five objects (apple, toothbrush, train, cubelarge and phone) from the GRAB training dataset for evaluation. 
The quantitative results and qualitative results are shown in \cref{tab:experimental_unseen} and \cref{fig:unknownobj}. 
As shown in \cref{tab:experimental_unseen}, we find that our method performs better on metrics for unseen objects compared to the metrics of unseen human identities in the paper. (The experiment in the main paper can be considered as a comparison with unseen human identities). 
From \cref{fig:unknownobj}, it is evident that our grasping performance is significantly better than that of OMOMO-V3.

\begin{table*}[!hbpt]

    \centering
    \fontsize{9}{11}\selectfont
 \begin{tabular}{l|l|cccccccc}
    
\toprule
 & Method	&	Hands JPE$\downarrow$	&	 MPJPE$\downarrow$	&	 MPVPE$\downarrow$	&	 FS$\downarrow$	&	 Coll. \%$\downarrow$	&	 C depth $\downarrow$ 	&	 F1$\uparrow$ 	&	 Cont dist$\downarrow$ \\
\midrule											
 \multirow{2}{*}{Unseen Objects} & OMOMO-V3	& 	 250.69 	&	139.17 	&	107.66 	&	\textbf{1.09} 	&	0.0000 	&	0.0000	&	 0.2923	&	 0.12	\\		

 & \textbf{DiffGrasp}	& 	\textbf{155.41} 	&	\textbf{84.25} 	&	\textbf{66.75} 	&	2.01 	&	0.0000 	&	0.0000 	&	\textbf{0.9233} 	&	\textbf{0.03} 	\\	
\midrule
\multirow{2}{*}{\parbox{3cm}{Unseen Human}} & OMOMO-V3	& 	327.16 	&	174.45 	&	137.52 	&	\textbf{1.03} 	&	0.0004 	&	0.0001 	&	0.1028 	&	0.15 \\

& \textbf{DiffGrasp}	& 	\textbf{209.85} 	&	\textbf{122.44} 	&	\textbf{100.88} 	&	2.22 	&	0.0023 	&	0.0001 	&	\textbf{0.7840} 	&	\textbf{0.04} 	\\	
\bottomrule			

    \end{tabular}

    \caption{\textit{Comparison }on GRAB \cite{GRAB:2020} dataset with unseen objects and unseen human identities. The experimental results of unseen human identities are from the Table 1 in the main paper.}
    
    \label{tab:experimental_unseen}
\end{table*}

\section{More Qualitative Results}
\label{sec:more results}

\begin{figure*}[!hbpt]
    \centering
    \includegraphics[width=0.73\textwidth]{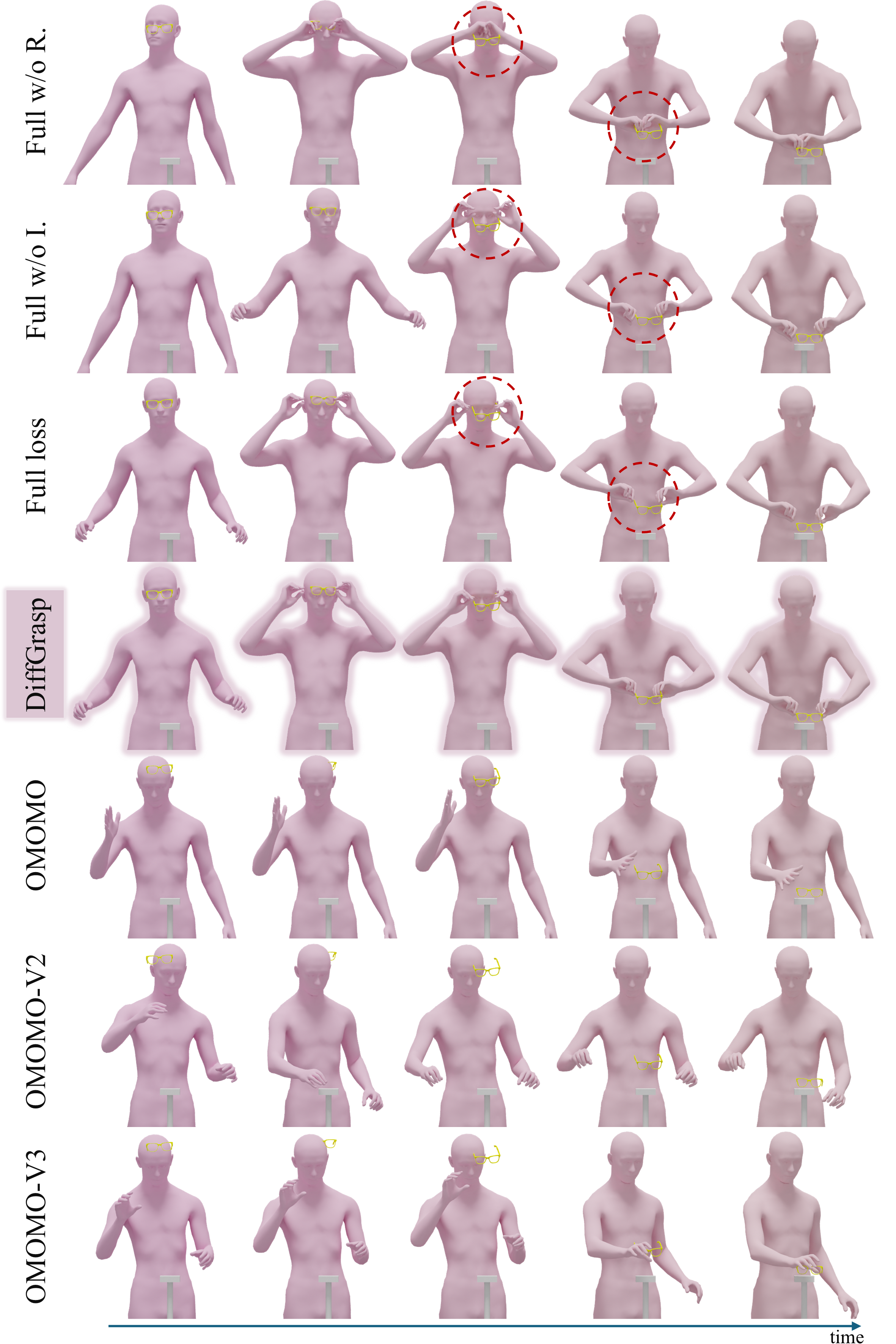}
    \caption{Comparison of the ability of different methods to jointly model the movement of different body parts. Our model (DiffGrasp) can generate more realistic results. As shown by the red dashed circle in the figure: in the "Full w/o R." row, we can see that the interaction loss brings the generated hand closer to the object's center, but the grasping pose is less natural; in the "Full w/o I." row, the reconstruction loss makes the generated results more natural, but the ability to perceive the object's position is weaker; in the "Full loss" row, we observe that the generated results combine the advantages of both losses, effectively perceiving the object's position while producing a more natural grasping pose. In contrast, OMOMO produces poor results, with generated poses that are dull and averaged, particularly in the case of OMOMO-V2. }
    \label{fig:more1}
\end{figure*}

\begin{figure*}[!hbpt]
    \centering
    \includegraphics[width=0.73\textwidth]{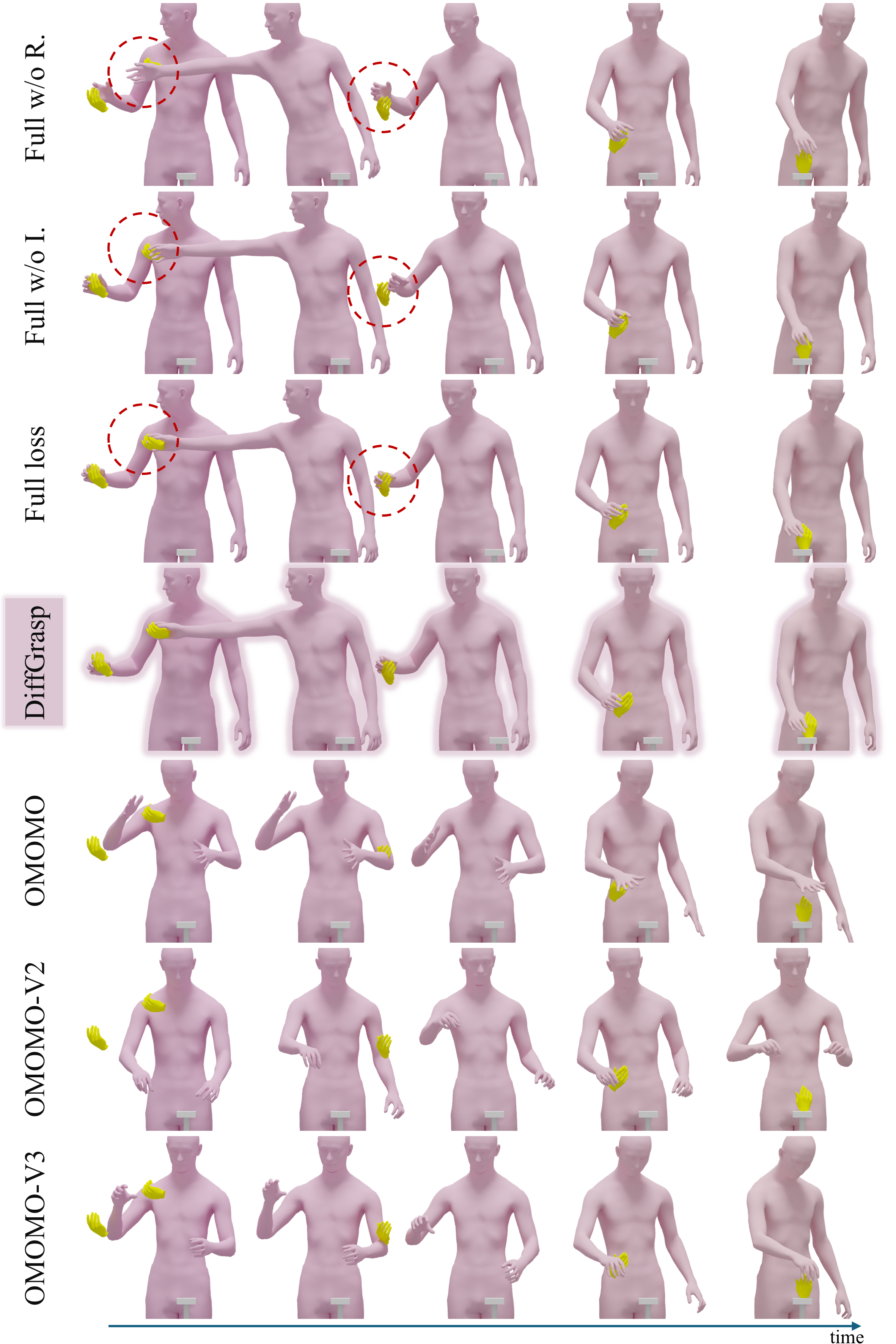}
    \caption{Comparison of the ability of different methods to adapt to large movements of objects. Our model (DiffGrasp) can generate more realistic results. The results in this figure are similar to those in Figure S6, highlighting the importance of our proposed two contact-aware loss terms. Additionally, the comparison shows that OMOMOs struggles to grasp objects that are out of reach.}
    \label{fig:more2}
\end{figure*}

\cref{fig:more1} illustrates the ability of different methods to jointly model all body parts in response to object motion, 
and \cref{fig:more2} shows the generated results of different methods for large-scale and rapid object movements. The results demonstrate that DiffGrasp can generate the most natural results. 
Moreover, the contact-aware interaction loss (Full w/o R.) demonstrates its ability to guide the modeling of hand positions relative to the object's spatial position. 
The contact-aware reconstruction loss (Full w/o I.) exhibits its capacity to produce more reasonable grasping poses. In contrast, all OMOMOs, especially when the object moves out of the human's immediate reach, fail to generate natural results that align well with the object's motion.

\section{Simplified Input Conditions}
\label{sec:simp condition}

In practical applications, having users specify the rotation matrix and translation for each frame can be labor intensive. To simplify our input requirements, we designed a method that uses specified rotation matrices and translation vectors for keyframes, with interpolated input for the intermediate frames. This can not only reduce the input burden, but also further evaluate the generalizability of our method to different input conditions. 

\subsubsection{Experiment Settings. }The objects used in this experiment are unseen objects. In addition to the initial and final positions of the object, we selected 4 additional in-between keyframes, dividing the motion sequence into 5 segments, with intermediate interpolation using simple quaternion interpolation. To better showcase our results, please refer to 3:10-3:29 at the end of the video.

We find that DiffGrasp adapts well to sparse key object motion frames as inputs. It still generates natural grasping results for the three specified conditions: large-range object movements, small-range object movements, and movements that involve interaction with other body parts.

\end{document}